\documentclass{article}
\PassOptionsToPackage{numbers, compress}{natbib}
\pdfoutput=1
\usepackage[final]{neurips_data_2024}

\usepackage{comment}
\usepackage[utf8]{inputenc} % allow utf-8 input
\usepackage[T1]{fontenc}    % use 8-bit T1 fonts
\usepackage{url}            % simple URL typesetting
\usepackage{booktabs}       % professional-quality tables
\usepackage{amsfonts}       % blackboard math symbols
\usepackage{nicefrac}       % compact symbols for 1/2, etc.
\usepackage{microtype}      % microtypography
\usepackage{xcolor}         % colors
\definecolor{cvprblue}{rgb}{0.21,0.49,0.74}
\usepackage[pagebackref,breaklinks,colorlinks,citecolor=cvprblue]{hyperref}
\usepackage{color, colortbl}
\usepackage{enumitem}
\usepackage[pdftex]{graphicx}
\usepackage{marvosym}
\usepackage{subcaption}
\usepackage{multirow}
\usepackage{amsmath}
\usepackage{wrapfig}
\usepackage{titlesec}
\usepackage{titletoc}
\usepackage{listings}
\usepackage{xcolor}
\usepackage[T1]{fontenc}

\newcommand{\methodname}{DrivingDojo}

\title{\methodname{} Dataset: Advancing Interactive and Knowledge-Enriched Driving World Model}

\author{%
  Yuqi Wang$^{1,2\dagger}$\thanks{Work done during an internship at Meituan. $\dagger$ equal contributions. $\textsuperscript{\Letter}$ Corresponding author}\quad 
  Ke Cheng$^{3\dagger}$ \quad 
  Jiawei He$^{1,2\dagger}$\quad 
  Qitai Wang$^{1,2\dagger}$\quad \and
  \textbf{Hengchen Dai}$^{3}$ \quad
  \textbf{Yuntao Chen}$^{4}\textsuperscript{\Letter}$\quad 
  \textbf{Fei Xia}$^{3}$ \quad
  \textbf{Zhaoxiang Zhang}$^{1,2,4}$ \quad \\[2mm]
  $^1$ New Laboratory of Pattern Recognition, Institute of Automation, Chinese Academy of Sciences \\
    $^2$ School of Artificial Intelligence, University of Chinese Academy of Sciences \\
    $^3$ Meituan Inc. \quad
    $^4$ Centre for Artificial Intelligence and Robotics, HKISI, CAS \\[1.5mm]
    Project page: \url{https://drivingdojo.github.io}
    \vspace{-10mm}
}

\begin{document}

\maketitle

\begin{figure}[htbp]
  \centering
\begin{subfigure}{0.49\linewidth}
    \includegraphics[width=1.0\linewidth]{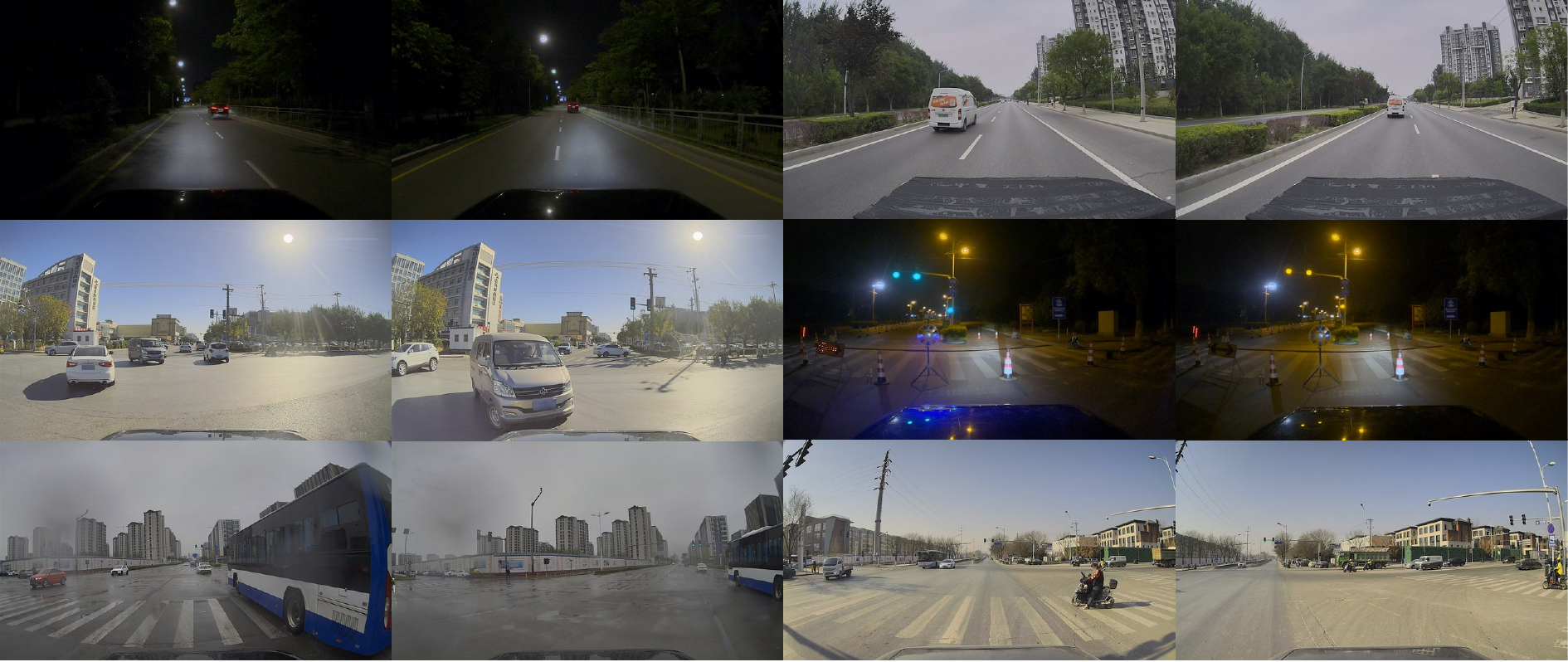}
    \vspace{-5mm}
    \caption{Rich ego actions.}
     \label{fig:actions}
    \end{subfigure}
    \begin{subfigure}{0.49\linewidth}
    \includegraphics[width=1.0\linewidth]{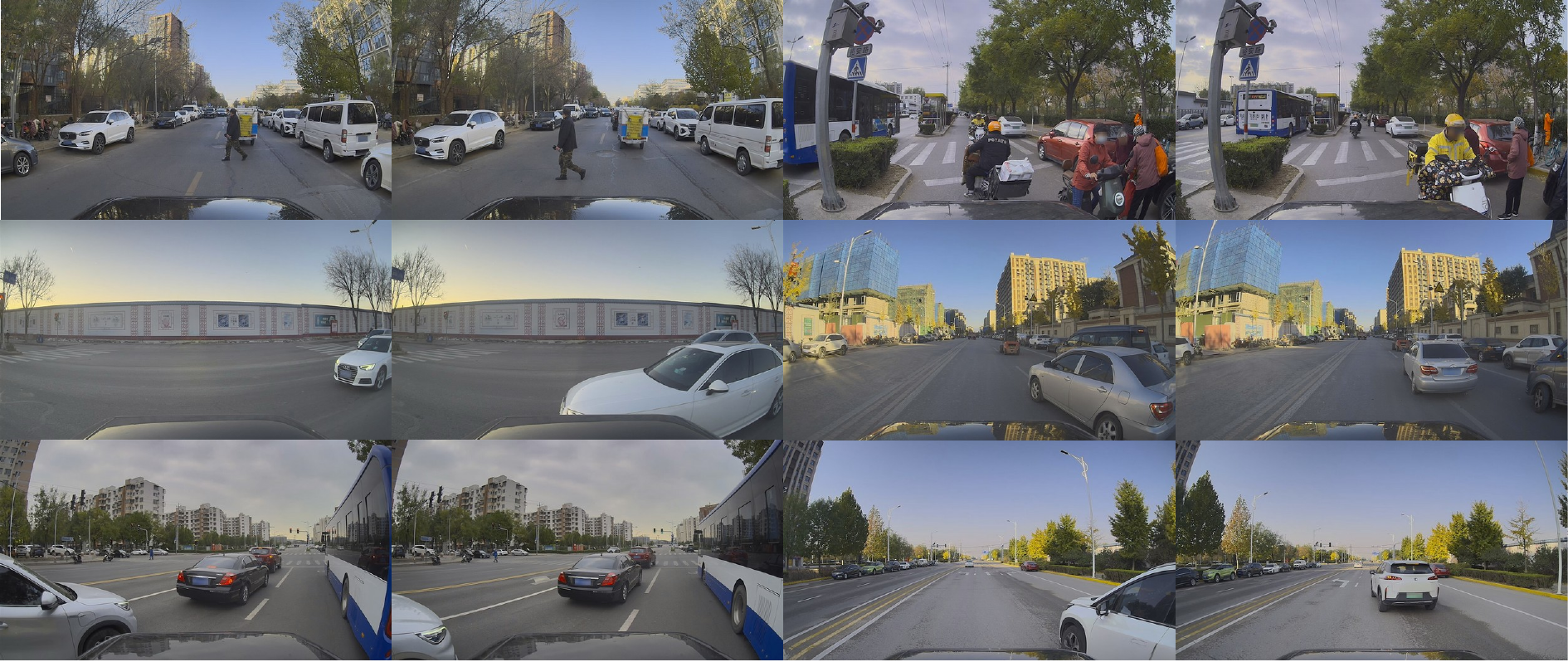}
    \vspace{-5mm}
    \caption{Multi-agent interplay.}
    \label{fig:interaction}
    \end{subfigure}
    \begin{subfigure}{0.49\linewidth}
    \includegraphics[width=1.0\linewidth]{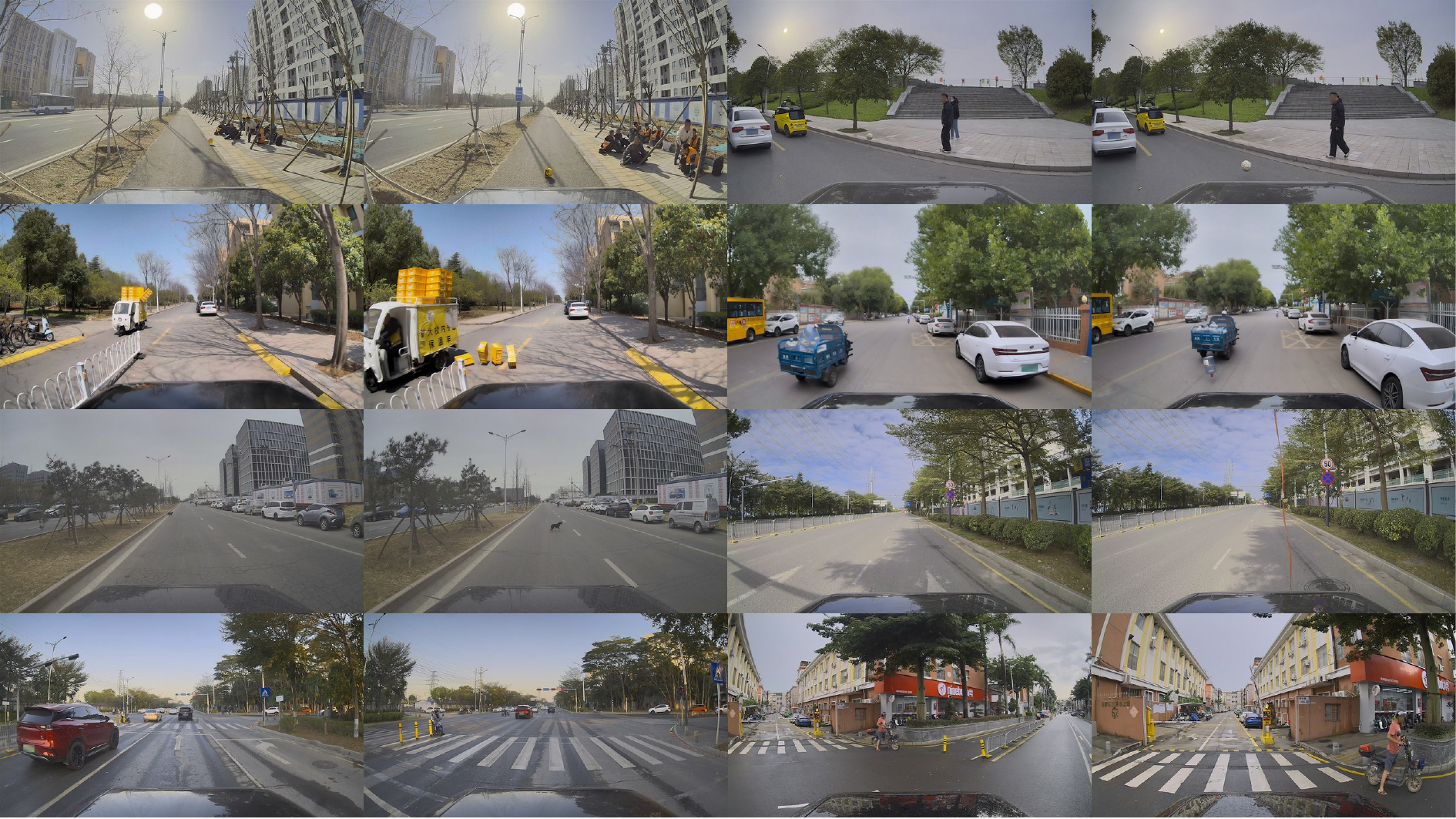}
    \vspace{-5mm}
    \caption{Rich open-world knowledge.}
  \label{fig:ood}
    \end{subfigure}
    \begin{subfigure}{0.49\linewidth}
    \includegraphics[width=1.0\linewidth]{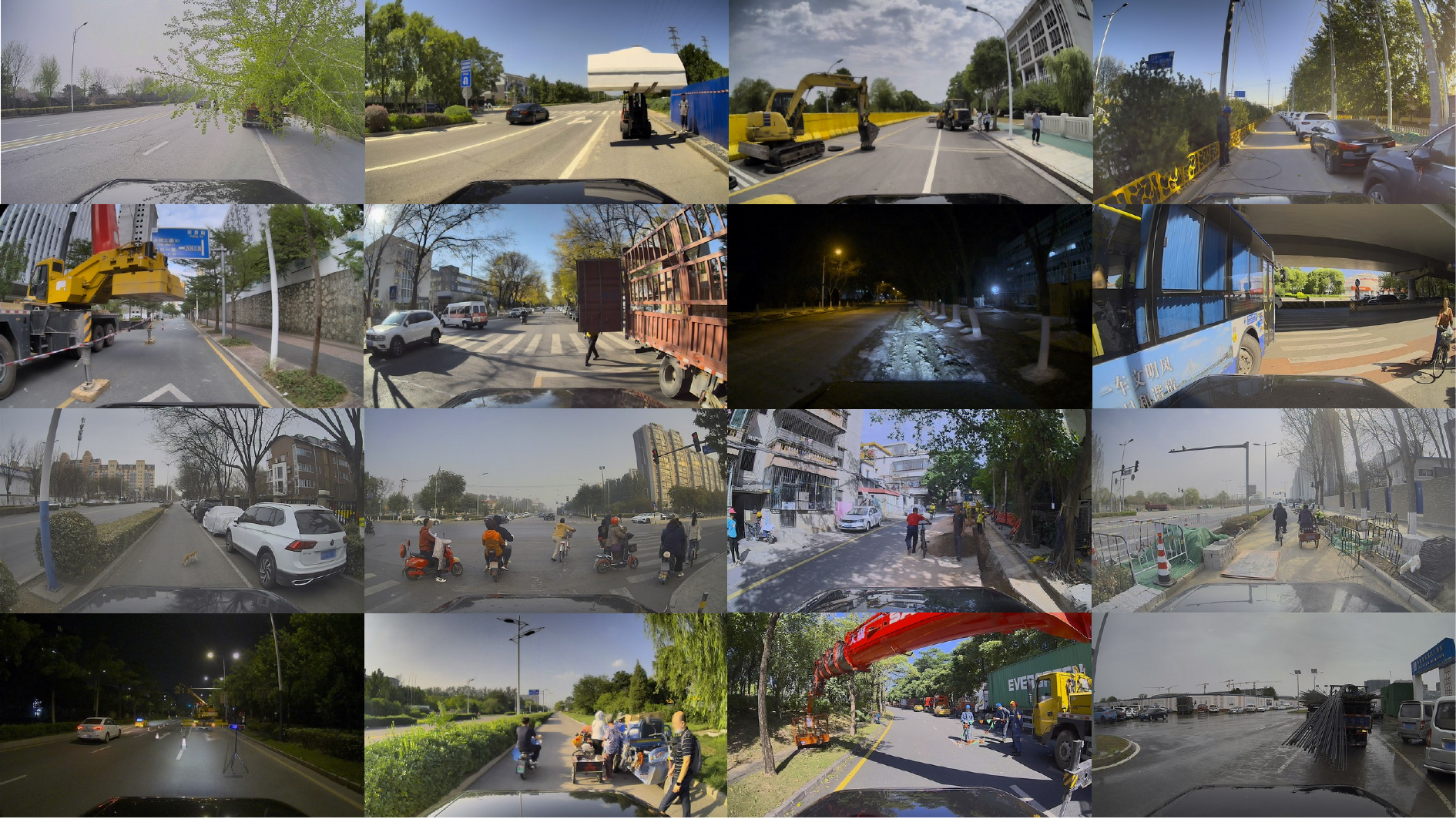}
    \vspace{-5mm}
    \caption{Diverse and rare cases.}
    \label{fig:diverse}
    \end{subfigure}
\caption{\textbf{Examples on \methodname{}.}
(a) showcases various driving actions, such as lane changes, abrupt braking at traffic control, and turning at intersections. (b) illustrates the ego-car's interactions with other dynamic agents, including cutting-in and cutting-off maneuvers. (c) displays encounters with rolling or falling objects, moving or floating unknown objects, and interactions with traffic lights and boom barriers. (d) presents diverse cases encountered in real-world driving scenarios.
}
\label{fig:all_example}
\vspace{-2mm}
\end{figure}

\begin{abstract}
Driving world models have gained increasing attention due to their ability to model complex physical dynamics. 
However, their superb modeling capability is yet to be fully unleashed due to the limited video diversity in current driving datasets. We introduce \methodname{}, the first dataset tailor-made for training interactive world models with complex driving dynamics. 
Our dataset features video clips with a complete set of driving maneuvers, diverse multi-agent interplay, and rich open-world driving knowledge, laying a stepping stone for future world model development.
We further define an action instruction following (AIF) benchmark for world models and demonstrate the superiority of the proposed dataset for generating action-controlled future predictions.
\end{abstract}

\begin{figure}[htbp]
  \centering
  \includegraphics[width=0.95\textwidth]{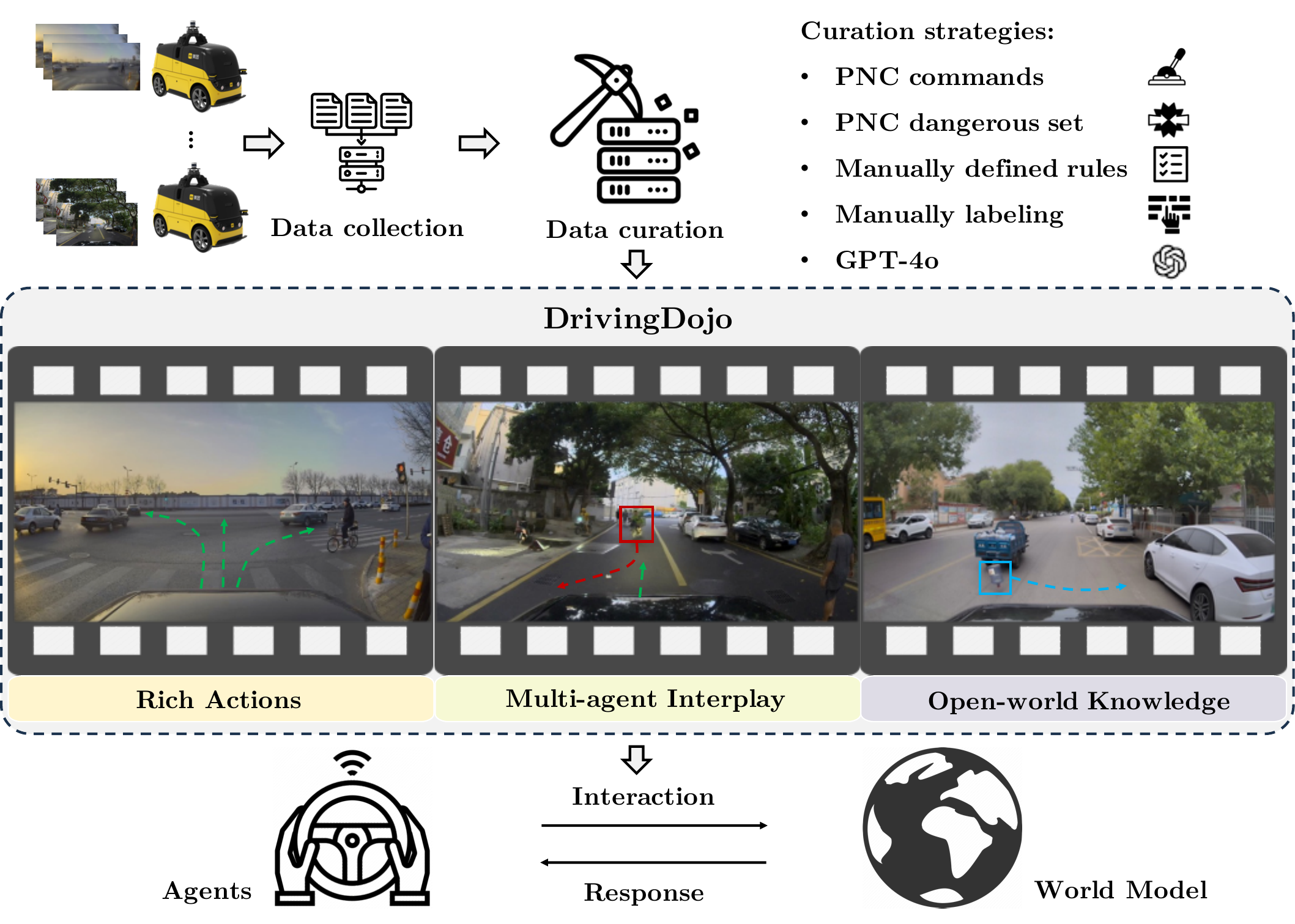}
    \vspace{-2mm}
  \caption{\textbf{Enhancing interactive and knowledge-enriched learning of world models.} Data plays a crucial role in modeling the world. \methodname{} is a large-scale video dataset curated from millions of daily collected videos, designed to investigate real-world visual interactions. \methodname{} features comprehensive actions, multi-agent interplay, and rich open-world driving knowledge, serving as a superb platform for studying driving world models.}
  \label{fig:motivation}
  \vspace{-2mm}
\end{figure}
\section{Introduction}
\label{sec:intro}

% 1. world model and development
% world model 
World models~\cite{ha2018world,hafner2020mastering,lecun2022path,hafner2023mastering} have gained increasing attention due to their ability to model complex real-world physical dynamics. They also hold potential as general-purpose simulators, capable of predicting future states in response to diverse action instructions.
Facilitated by advancements in video generation techniques~\cite{yan2021videogpt,ho2022imagen,blattmann2023align,blattmann2023stable}, models like Sora have achieved remarkable success in producing high-quality videos, thereby opening up a new avenue that treats video generation as real-world dynamics modeling problem~\cite{vondrick2016generating, hafner2019learning, yang2024position}. 
Generative world models, in particular, hold significant promise as real-world simulators and have garnered extensive research in the field of autonomous driving~\cite{hu2023gaia,wang2023drivedreamer,jia2023adriver,wang2024driving,yang2024generalized,zhao2024drivedreamer, gao2024vista}.

% 2. Existing world model problems
However, existing driving world models fall short of meeting the requirements of model-based planning in autonomous driving, which aims to improve driving safety in scenarios with diverse ego maneuvers and intricate interaction between the ego vehicle and other road users.
These models perform well for non-interactive in-lane maneuvers but have shown limited capability in following more challenging action instructions like lane change.
One significant roadblock to building next-generation driving world models lies in the datasets. 
Autonomous driving datasets commonly used in current world model literature like nuScenes~\cite{caesar2020nuscenes}, Waymo~\cite{sun2020scalability}, and ONCE~\cite{mao2021one}, are primarily designed and curated in a perception-oriented manner. 
As a result, it contains limited driving patterns and multi-agent interactions, which may not fully capture the complexities of real-world driving scenarios.
The scarcity of interaction data limits the ability of models to accurately simulate and predict the complex dynamics of real-world driving environments.

% 3. datasets
In this paper, we propose \textbf{\methodname{}}, a large-scale driving video dataset designed to simulate real-world visual interaction. 
As illustrated in Figure~\ref{fig:all_example}, \methodname{} features action completeness, multi-agent interplay, and open-world driving knowledge. 
Our dataset aims to unleash the full potential of world models in action instruction following by including rich longitudinal maneuvers like acceleration, emergency braking and stop-and-go as well as lateral ones like U-turn, overtaking, and lane change.
Besides, we explicitly curate the dataset to include a large volume of trajectories containing multi-agent interplays like cut-in, cut-off, and head-to-head merging.
Finally, \methodname{} taps into the open-world driving knowledge by including videos containing rare events sampled from tens of millions of driving video clips, including crossing animals, falling bottles and debris.
As shown in Figure~\ref{fig:motivation}, we hope that \methodname{} could serve as a solid stepping stone for developing next-generation driving world models.

% 4. method & metric, challenges
To measure the progress of driving scene modeling, we propose a new action instruction following (AIF) benchmark to assess the ability of world models to perform plausible future rollouts.
The AIF benchmark measures the visual and structural fidelity of videos generated by world models in an action-conditioned manner.
We propose the AIF errors calculated on the withheld validation data to evaluate the long-term motion controllability for generated videos.
The error is defined as the mean error between the actions estimated from the generated video and the given action instructions.
Then the baseline world model is evaluated on our \methodname{} AIF benchmark, for in-domain data and out-of-domain images or action conditions.  

% 5. Contributions
Our major contributions are as follows. 
(1) We design a large-scale driving video dataset to facilitate research in world model for autonomous driving. 
Compared to previous datasets in Table~\ref{tab:compare}, our dataset features complete driving actions, diverse multi-agent interplay, and rich open-world driving knowledge.
(2) We design an action instruction following task for driving world model and provide corresponding video world model baseline methods.
(3) Benchmark results on both driving video generation and action instruction following show that there are plenty of new opportunities for future driving world model development on our new dataset.
\begin{table}[t]
\centering

\caption{\textbf{A comparison of driving datasets for world model}. This comparison emphasizes the diversity of the video content, placing less focus on annotations or sensor data. $^*$ denotes that the videos are curated from our data pool of around 7500 hours.}
\label{tab:compare}
\resizebox{\textwidth}{!}{
\begin{tabular}{c|cccccc}
\toprule
Dataset & Videos & \multirow{2}{*}{\shortstack{Duration\\(hours)}} & \multirow{2}{*}{\shortstack{Ego\\Trajectory}} & \multirow{2}{*}{\shortstack{Complete\\Actions}} & \multirow{2}{*}{\shortstack{Multi-agent\\Interplay}} & \multirow{2}{*}{\shortstack{Open-world\\Knowledge}} \\
\\
\midrule
nuScenes~\cite{caesar2020nuscenes} & 1k & 5.5 & \checkmark & & \\ 
Waymo~\cite{sun2020scalability} & 1k & 11& \checkmark & &\\
OpenDV-2k~\cite{yang2024generalized} & 2k& 2059 & &\checkmark & \\
nuPlan~\cite{caesar2021nuplan} & - & 1500& \checkmark & \checkmark& \checkmark&\\
\midrule
\methodname{} (Ours) & 18k &  150$^*$&\checkmark & \checkmark & \checkmark & \checkmark \\
\bottomrule
\end{tabular}
}
\vspace{-5mm}
\end{table}

\section{Related Works}
\subsection{Autonomous Driving Datasets}

\vspace{-2mm}
\paragraph{Datasets for perception.}
The driving dataset has played a crucial role in advancing computer vision in recent years, aiming to achieve comprehensive perception and understanding surrounding the ego vehicle. Initially, perception in autonomous driving relied on 2D image-based perception. Datasets like Cityscapes~\cite{cordts2016cityscapes}, Mapillary Vistas~\cite{neuhold2017mapillary}, and BDD100k~\cite{yu2020bdd100k} provided instance-level masks for learning tasks. With the integration of LiDAR sensors and advancements in 3D perception, datasets like KITTI~\cite{geiger2012we}, nuScenes~\cite{caesar2020nuscenes}, and Waymo~\cite{sun2020scalability} have emerged as standard benchmarks for various 3D perception tasks. Additionally, datasets like ONCE~\cite{mao2021one}, Argoverse~\cite{chang2019argoverse,wilson2023argoverse}, and others~\cite{huang2019apolloscape,geyer2020a2d2,alibeigi2023zenseact} are also utilized for studying various perception tasks.
\vspace{-3mm}
\paragraph{Datasets for prediction and planning.}
In recent years, there's been increasing attention on prediction and planning in autonomous driving. Prediction involves anticipating the behavior of other agents, while planning relates to the behavior of the ego vehicle.
Prediction methods typically rely on semantic maps and dynamic traffic light statuses to anticipate future vehicle motions. Notable datasets in this area include Argoverse Motion Forecasting~\cite{chang2019argoverse}, Waymo Open Motion Dataset~\cite{ettinger2021large}, Lyft Level 5 Prediction Dataset~\cite{houston2021one}, and nuScenes Prediction~\cite{caesar2020nuscenes} challenge. Additionally, the Interaction dataset~\cite{zhan2019interaction} provides interactive driving scenarios with semantic maps derived from drones and traffic cameras, enriching the understanding of complex driving interactions.
Transitioning to planning, CARLA~\cite{dosovitskiy2017carla} stands out as an open-source simulator designed to simulate real-world traffic scenarios, providing a platform for testing and validating planning algorithms. Complementing this, nuPlan~\cite{caesar2021nuplan} introduces the first closed-loop planning benchmark for autonomous vehicles, closely mirroring real-world scenarios.

\subsection{World Model}
\vspace{-2mm}
\paragraph{Learning world models.}
World models~\cite{ha2018world, lecun2022path} enable next-frame prediction based on action inputs, aiming to build general simulators of the physical world. However, learning dynamic modeling in pixel space is challenging, leading previous image-based world models to focus on simplistic gaming environments or simulations~\cite{hafner2019dream,hafner2020mastering, chen2021transdreamer,wu2022slotformer,seo2022reinforcement,seo2023masked,hafner2023mastering}.
With advances in video generation, models like Sora can now produce high-definition videos up to one minute long with natural, coherent dynamics. This progress has encouraged researchers to explore world models in real-world scenarios. 
DayDreamer~\cite{wu2023daydreamer} applies the Dreamer algorithm to four robots, allowing them to learn online and directly in the real world without simulators, demonstrating that world models can facilitate faster learning on physical robots.
Genie~\cite{bruce2024genie} demonstrates interactive generation capabilities using vast internet gaming videos and shows potential for robotics applications. UniSim~\cite{yang2023learning} aims to create a universal simulator for real-world interactions using generative modeling, with applications extending to real-robot executions.

\vspace{-3mm}
\paragraph{World model for autonomous driving.}
World models serving as real-world simulators have garnered widespread attention~\cite{guan2024world, zhu2024sora} and can be categorized into two main branches. 
The first branch explores agent policies in virtual simulators. MILE~\cite{hu2022model} employed imitation learning to jointly learn the dynamics model and driving behavior in CARLA~\cite{dosovitskiy2017carla}. Think2Drive~\cite{li2024think2drive} proposed a model-based RL method in CARLA v2, using a world model to learn environment transitions and acting as a neural simulator to train the planner.
The second branch focuses on simulating and generating real-world driving scenarios. GAIA-1~\cite{hu2023gaia} introduced a generative world model for autonomous driving, capable of simulating realistic driving videos from inputs like images, texts, and actions. DriveDreamer~\cite{wang2023drivedreamer} emphasized scenario generation, leveraging HD maps and 3D boxes to enhance video quality. Drive-WM~\cite{wang2024driving} was the first to propose a multiview world model for generating high-quality, controllable multiview videos, exploring applications in end-to-end planning.
ADriver-I~\cite{jia2023adriver} constructed a general world model based on MLLM and diffusion models, using vision-action pairs to auto-regressively predict current frame control signals. DriveDreamer2~\cite{zhao2024drivedreamer} leveraged LLMs and text prompts to generate diverse driving videos in a user-friendly manner. Unlike previous methods that focused on model design, OpenDV-2K~\cite{yang2024generalized} addressed the issue of training data by collecting over 2000 hours of driving videos from the internet.
Previous research has predominantly addressed static scene generation, with limited emphasis on multi-agent interplays. Our dataset enables the exploration of world model predictions within dynamic, interactive driving scenarios.

\begin{figure}[htbp]
  \centering
   \begin{subfigure}{0.45\linewidth}
  \includegraphics[width=\linewidth,trim=0 0 0 0, clip]{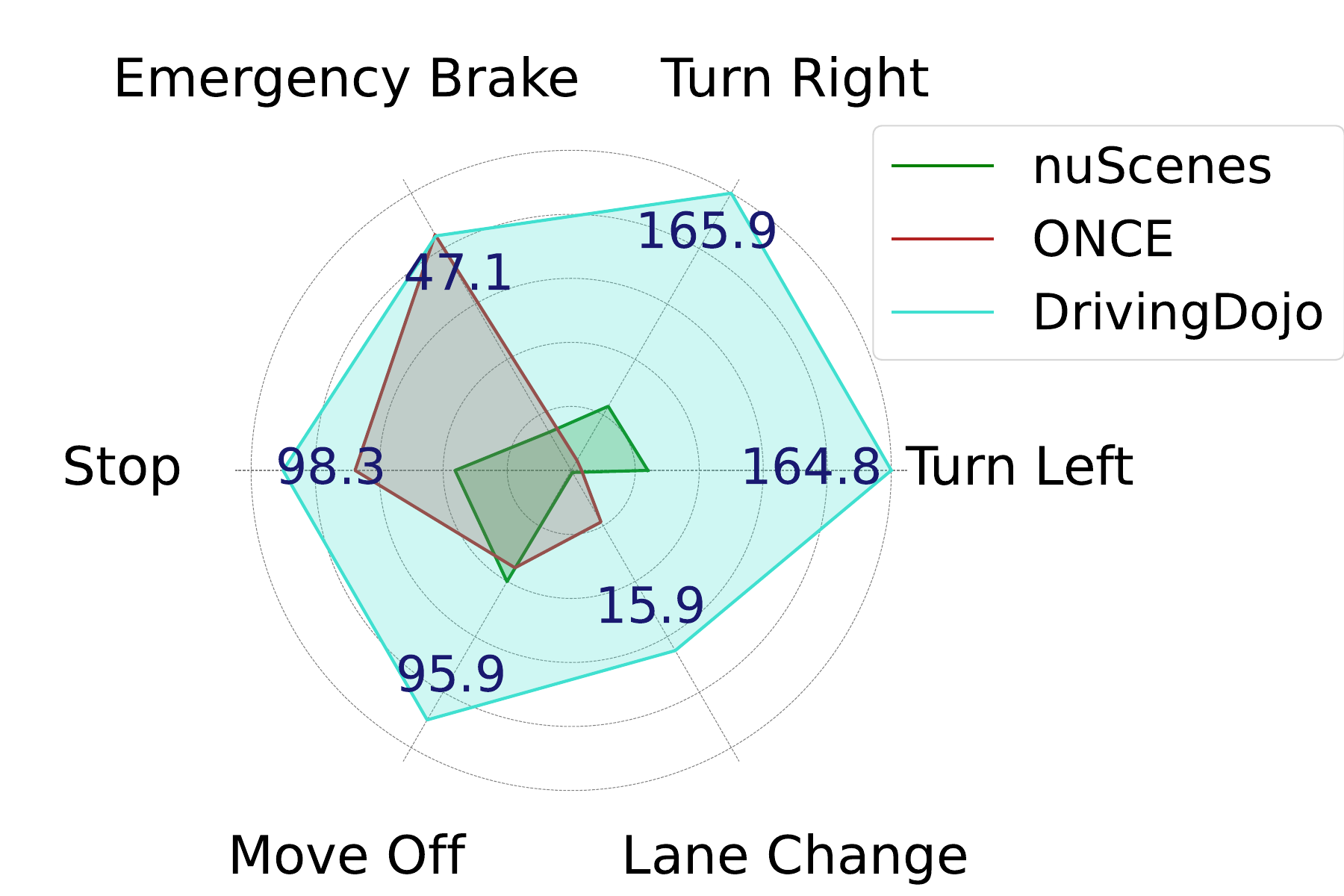} 
    \caption{Action distribution}
      \label{fig:act_dist}
    \end{subfigure}
    \begin{subfigure}{0.45\linewidth}
    \includegraphics[width=\linewidth]{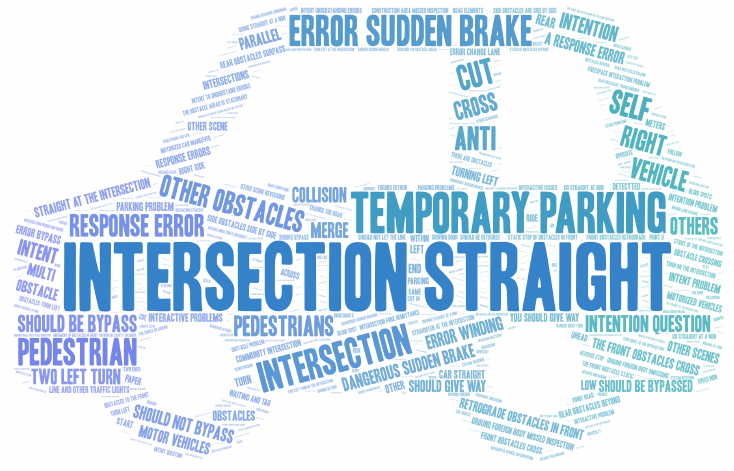}
    \caption{Video clip description}
    \label{fig:video_cloud}
    \end{subfigure}
  \caption{\textbf{The strengths of the \methodname{} dataset.} (a) illustrates a comparison of action distributions among nuScenes, ONCE, and our \methodname{}. We compare the average hourly event counts of driving actions. (b) presents the distribution of text descriptions for the video clips in \methodname{}.}
  \vspace{-5mm}
\end{figure}
\section{The \methodname{} Dataset}
% motivation
Our goal is to provide a large and diverse action-instructed driving video dataset \methodname{} to support the development of driving world models.
To accomplish this, we extract highly informative clips from a video pool collected through fleet data, spanning several years and comprising more than 500 operating vehicles across multiple major Chinese cities.
As a result, our \methodname{} features diverse ego actions, rich interactions with road users, and rare driving knowledge which are crucial for high-quality future forecasting as shown in Table~\ref{tab:summary}.

We begin with the design principles of \methodname{} and its uniqueness compared with existing datasets in Section~\ref{sec:action}-~\ref{sec:knowledge}.
We then describe the data curation procedure and statistics in Section~\ref{sec:data}. 
Here, we only describe the design principles. More detailed information refer to the Appendix.

\begin{table}[htb]
\centering
\caption{\textbf{\methodname{} dataset constitution.} The dataset is organized into three subsets: \methodname{}-Action, \methodname{}-Interplay, and \methodname{}-Open, to support research on specific tasks.}
\label{tab:summary}
\resizebox{\textwidth}{!}{
\begin{tabular}{c|c|c|ccc}
\toprule
\textbf{Dataset} & \textbf{Videos} & \textbf{Type} &\textbf{Camera}& \textbf{Ego Trajectory}& \textbf{Text Description} \\
\midrule
\methodname{} & 17.8k & total & \checkmark & \checkmark & \checkmark \\ \midrule
\methodname{}-Action & 7.9k & rich ego-actions &  \checkmark & \checkmark & \\
\methodname{}-Interplay & 6.2k & multi-agent interplay &  \checkmark & \checkmark\\
\methodname{}-Open & 3.7k & open-world knowledge& \checkmark &  \checkmark & \checkmark \\
\bottomrule
\end{tabular}}
\end{table}

\subsection{Action Completeness}
\label{sec:action}

%motivation
Using the driving world model as a real-world simulator requires it to follow action prompts accurately.
Existing autonomous driving datasets, such as ONCE~\cite{mao2021one} and nuScenes~\cite{caesar2020nuscenes}, are generally curated for developing perception algorithms and thus lack diverse driving maneuvering.

To enable the world model to generate an infinite number of high-fidelity, action-controllable virtual driving environments, we create a subset called \methodname{}-Action that features a balanced distribution of driving maneuvers. 
This subset includes a diverse range of both longitudinal maneuvers, such as acceleration, deceleration, emergency braking, and stop-and-go driving, as well as lateral maneuvers, including lane-changing and lane-keeping. 
As demonstrated in Figure~\ref{fig:act_dist}, our \methodname{}-Action subset offers a significantly more balanced and complete set of ego actions compared to existing autonomous driving datasets.

\subsection{Multi-agent Interplay}
\label{sec:interplay}
Besides navigating in a static road network environment, modeling the dynamics of multi-agent interplay like merge and yield is also a crucial task for world models.
However, current datasets are either built without considering multi-agent interplays, such as nuScenes~\cite{caesar2020nuscenes} and Waymo~\cite{sun2020scalability}, or are constructed from large-scale internet videos that lack proper curation and balancing, like OpenDV-2K~\cite{yang2024generalized}.

To address this issue, we design the \methodname{}-Interplay subset focusing on interactions with dynamic agents as a core component of the dataset. 
As shown in Figure~\ref{fig:interaction}, we curate this subset to include at least one of the following driving scenarios: cutting in/off, meeting, blocked, overtaking, and being overtaken. 
These scenarios encompass a variety of realistic situations, such as vehicles cutting into lanes, encounters with oncoming traffic, and the necessity for emergency braking. 
By incorporating these diverse scenarios, our dataset enables world models to better understand and anticipate complex interactions with dynamic agents, thereby improving their performance in real-world driving conditions.

\subsection{Rich Open-world Knowledge}
\label{sec:knowledge}
In contrast to perception and prediction models, which compress high-dimensional sensor input into low-dimensional vector representations, world models exhibit a superior modeling capacity by operating in the pixel space. 
This increased capacity enables world models to effectively capture the intricate dynamics of open-world driving scenarios, such as animals unexpectedly crossing the road or parcels falling off the trunks of vehicles.

However, existing datasets, either perception-oriented ONCE~\cite{mao2021one} or planning-oriented ones like nuPlan~\cite{caesar2021nuplan}, do not have adequate data for developing and assessing the long-tail knowledge modeling ability of world models.
Therefore, we place a unique emphasis on including rich open-world knowledge video clips and construct the \methodname{}-Open subset.
As shown in Figure~\ref{fig:ood}, describing open-world driving knowledge like this is challenging due to its complexity and variability, but these scenarios are crucial for ensuring safe driving.

The \methodname{}-Open subset consists of 3.7k video clips about the open-world knowledge in driving scenarios. 
This subset is curated from fleet data that includes unusual weather, foreign objects on the road surface, floating obstacles, falling objects, taking over cases, and interactions with traffic lights and boom barriers. 
A word cloud of video descriptions for \methodname{}-Open are shown in Figure~\ref{fig:video_cloud}. 
\methodname{}-Open serves as an invaluable supplementary for driving world modeling by including driving knowledge beyond simply interacting with structured road networks and other regular road users.

%-------------------------------------------------------------------------
\begin{figure}[htbp]
  \centering
  \includegraphics[width=1.0\textwidth]{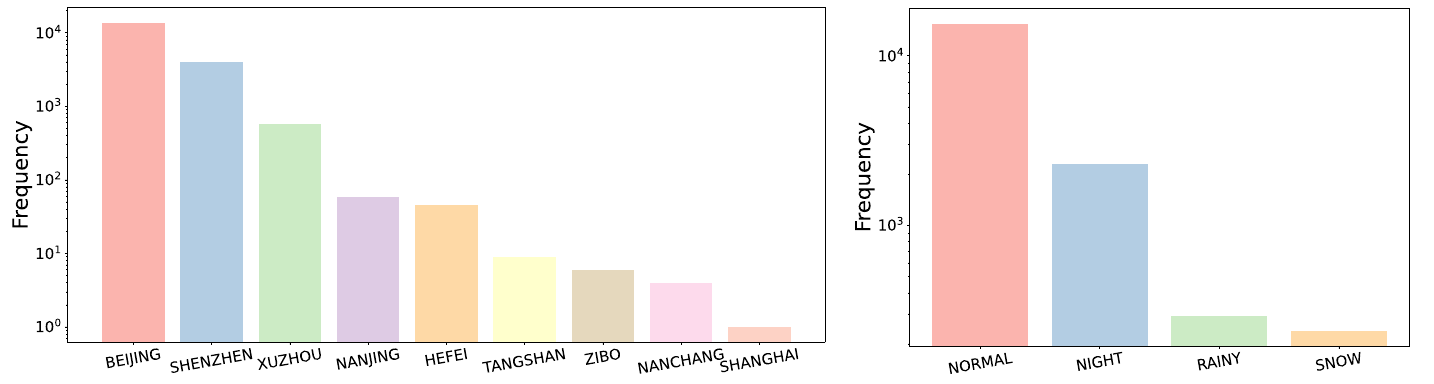}
  \caption{\textbf{Descriptive statistics of the \methodname{} dataset.} The dataset was collected from various regions across China, including nighttime and rainy/snowy conditions.}
  \label{fig:distribution}
  \vspace{-5mm}
\end{figure}

\subsection{Data Curation and Statistics}
\label{sec:data}
\noindent\textbf{Dataset statistics.} The \methodname{} dataset contains around 18k videos with resolution of 1920$\times $1080 and frame rate at 5 fps.
Our video clips are collected from major Chinese cities including Beijing, Shenzhen, Xuzhou, etc., as shown in Figure~\ref{fig:distribution}.
Furthermore, these videos are recorded in diverse weather conditions at different daylight conditions. 
All videos are paired with synced camera poses derived from the HD-Map powered high precision localization stack onboard.
Videos in the \methodname{}-Open subset are paired with text descriptions about the rare event happening in each video. More details are in the Appendix.

\noindent\textbf{Data collection.}
We collected multi-modal fleet data using the platform of Meituan's autonomous delivery vehicles. 
Our dataset consists of video clips recorded by the front-view camera with a horizontal field of view of 120$^\circ$ to capture comprehensive visual information. 
The raw data is collected from multiple Chinese cities between May 2022 and May 2024, amassing a total of 900,000 videos and approximately 7,500 hours of driving footage pre-filtered before recording.

\noindent\textbf{Data curation.}
In order to ensure both the data diversity as well as balanced ego action and multi-agent interplay distribution, we include fleet data with different criteria.
The data sources of \methodname{} include 1) intervention data from safety inspectors during vehicle operation, 2) emergency brake data from automatic emergency braking, 3) randomly sampled 30-second general videos from collected videos, 4) selected distinct scenarios such as traffic light changes, barrier opening, left and right turns, straight crossings, vehicle encounters, lane changes, and pedestrian interactions, 5) manually sorted rare data containing moving and static foreign objects on the road, floating obstacles, falling and rolling objects.
The curation details are in the Appendix.

\noindent\textbf{Personal Identification Information (PII) removal.}
To avoid privacy infringement and obey the regulation laws, we employ a high precision license plate and face detectors~\cite{Jocher_Ultralytics_YOLO_2023} to detect and blur these PII for each frame of all videos. 
An in-house annotation team and the authors have manually double-checked that the PII removal procedure is correctly carried out for all the videos.

% ------------------------------------------------------------------------
\section{\methodname{} for World Model}
To facilitate the study of world models in autonomous driving, we define a novel action instruction following (AIF) task. 
We provide baseline methods (Section~\ref{sec:baseline}) and evaluation metrics (Section~\ref{sec:eval}), enabling further investigations. More details are described in the Appendix.

\subsection{Action Instruction Following}
\label{sec:task}
Action-controllable video forecasting is the core ability of world models~\cite{bruce2024genie}.
Instead of solely focusing on predicting high-quality video frames, action instruction following requires world models to take both the initial video frame and ego action prompts into consideration for predicting corresponding world responses.
Given the initial image \( I_t \) and a sequence of actions \( \{A_t,...,A_{t+k}\} \), the model \( f_{\theta} \) predicts future states \( \{I_{t+1},...,I_{t+k}\} \) as:
\begin{equation}
    \{I_{t},...,I_{t+k}\} = f_{\theta}(I_t, \{A_t,...,A_{t+k}\}).
\end{equation}
Here, \( \{A_t,...,A_{t+k}\} \) refers to the action prompts for each frame, with trajectories \( A_t = (\Delta x_t, \Delta y_t) \) in our experiment.
\( f_{\theta} \) represents the world model, and \( \{I_{t+1},...,I_{t+k}\} \) signifies the visual prediction for subsequent \( k \) frames.

\subsection{Model Architecture}
\label{sec:baseline}
We propose \methodname{} baseline, a video generation model based on Stable Video Diffusion (SVD)~\cite{blattmann2023stable}. While SVD is a latent diffusion model for image-to-video generation, we extend its capability to generate videos conditioned on action.
For the AIF task, we encode the value of each action sequence into a 1024-dimensional vector using a Multilayer Perceptron (MLP). Subsequently, the action feature is concatenated with the first-frame image feature and passed into the U-Net~\cite{ronneberger2015u}. 

% ------------------------------------------------------------------------
\subsection{Evaluation Metrics}
\label{sec:eval}
\noindent{\textbf{Visual quality.}}
To evaluate the quality of the generated video, we utilize
FID (Frechet Inception Distance)~\cite{heusel2017gans} and FVD (Frechet Video Distance)~\cite{unterthiner2018towards} as the main metrics.

\noindent{\textbf{Action instruction following.}}
We propose the action instruction following (AIF) errors $E^{\text{AIF}}_x$ and $E^{\text{AIF}}_y$ to measure the consistency between the generated video and the input action conditions.
Given the generated video sequences $\{I_{t},...,I_{t+k}\}$, we estimate vehicle trajectories in the generated videos with 
the offline visual structure-from-motion (SfM) implementation like COLMAP~\cite{schoenberger2016sfm,schoenberger2016mvs}: $\{\widetilde{A}_t,...,\widetilde{A}_{t+k}\} = \text{SfM}(\{I_{t},...,I_{t+k}\})$, where $\{\widetilde{A}_t,...,\widetilde{A}_{t+k}\}$ are estimated trajectories of unknown scale.
We estimated the scale factor $\hat{S}$ for the predicted trajectory by minimizing the error between estimated and input ego-motion in the first $N$ frames. 
We compare the estimated actions with the ground-truth action instructions $\{A_t,...,A_{t+k}\}$ and report the mean absolute error for both lateral ($E^{\text{AIF}}_y$) and longitudinal ($E^{\text{AIF}}_x$) actions:

\begin{equation}
     (E^{\text{AIF}}_x,E^{\text{AIF}}_y) = \frac{\sum\limits_{i=0}^{k}|{A}_{t+i}-\widetilde{A}_{t+i}*\hat{S}|}{k+1},
\end{equation}
where the scale factor $\hat{S} = \mathop{\arg\min}\limits_{S}\sum\limits_{i=0}^{N}|{A}_{t+i}-\widetilde{A}_{t+i}*S|$.

\begin{table}[htbp]
\centering
\caption{\textbf{Comparison of visual prediction fine-tuning across different datasets.}, $\dagger$ indicates using camera sweeps data. The performance is zero-shot evaluated on the OpenDV-2K dataset.}
\label{tab:exp_visual}
\begin{tabular}{c|c|c|c|c}
\toprule
\textbf{Method} & \textbf{Fine-tuning}& \textbf{Evaluation} &  \textbf{FID} & \textbf{FVD} \\
\midrule
\textcolor{gray}{SVD} & \textcolor{gray}{OpenDV-2K} & \textcolor{gray}{OpenDV-2K} & \textcolor{gray}{18.27} & \textcolor{gray}{321.05} \\
\midrule
SVD & - & OpenDV-2K&24.17& 580.94\\
SVD & nuScenes$\dagger$ & OpenDV-2K& 21.05 & 395.04 \\
SVD & \methodname{} & OpenDV-2K& \textbf{19.20}& \textbf{343.91}\\
\bottomrule
\end{tabular}
\vspace{-2mm}
\end{table}
\section{Experiments}

\subsection{Results of Visual Prediction}
To illustrate the richness of behaviors and dynamics within our dataset, we compare video fine-tuning quality across various datasets.
In Table~\ref{tab:exp_visual}, we random selected 256 video segments from the OpenDV-2K dataset~\cite{yang2024generalized} as our test set and evaluated fine-tuning performance of SVD~\cite{blattmann2023stable} model across various datasets. The results indicate that models trained on our dataset exhibit better visual quality.

\subsection{Results of Action Instruction Following}

\begin{figure}[htbp]
  \centering
  \includegraphics[width=1.0\textwidth]{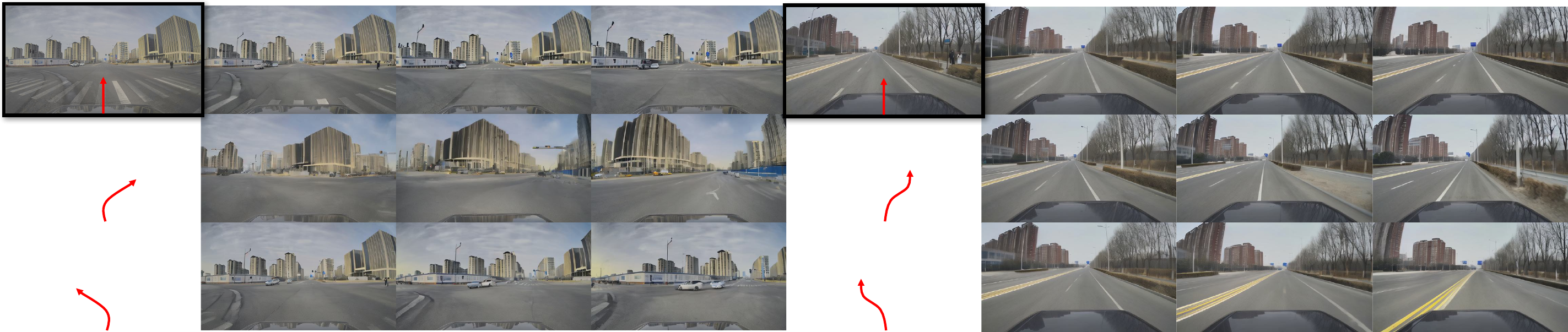} 
  \caption{\textbf{Predicting multiple futures based on different actions.} Left: going straight, turning left, and turning right at a crossing; Right: changing to the left lane, staying in the current lane, and changing to the right lane.}
  \label{fig:diverse_action}
\end{figure}

\noindent{\textbf{Diverse driving behaviors.}}
Based on different sequences of actions, our model is able to generate multiple possible futures.
As shown in Figure~\ref{fig:diverse_action}, we showcase the model's capability to execute forward, left turn, and right turn maneuvers at intersections, as well as lane-changing to the left or right, and maintaining on straight roads.

\noindent{\textbf{Action instruction following.}}
Although qualitative evaluations demonstrate the powerful generative ability of our model, we also endeavor to measure the accuracy of action instruction following quantitatively.
We seek to evaluate whether the video trajectories generated by the model closely adhere to our expected route paths. This serves as a fundamental assurance for the future application of world model.
As shown in Table~\ref{tab:exp_action}, with the in-domain actions (original action sequences of the test video) as conditions, videos generated by the baseline world model trained on \methodname{} exhibit strong loyalty towards the action instructions. The mean action error in each video frame is limited to only 10 cm in the lateral or longitudinal directions. In row 3, feeding the model with the same initial images and randomly sampled action instructions slightly increases the mean action errors. When the model is applied zero-shot to initial images from OpenDV-2K~\cite{yang2024generalized} and fed with randomly sampled action instructions, its generated videos still demonstrate considerable consistency to the action instructions. Note that the proposed action instruction following errors can sensitively reflect the impact of out-of-domain inputs on the performance of the model.

\begin{table}[htbp]
\centering
\caption{\textbf{Action instruction following on the \methodname{} dataset}. GT refers to using real images to test the accuracy of the reconstructed trajectory. $^*$  denotes the model is applied zero-shot to this dataset without fine-tuning.}
\label{tab:exp_action}
\begin{tabular}{c|c|cc|cc}
\toprule
Action Type & Test Dataset& FID & FVD & $E^{\text{AIF}}_x$($\downarrow$) & $E^{\text{AIF}}_y$($\downarrow$) \\
\midrule
\textcolor{gray}{In-Domain} & \textcolor{gray}{\methodname{}(GT)} &  \textcolor{gray}{-} & \textcolor{gray}{-} & \textcolor{gray}{0.036m} &	\textcolor{gray}{0.019m} \\
\midrule
In-Domain & \methodname{} & 37.07 & 658.72 & 0.100m & 0.062m \\
Out-of-Domain & \methodname{}& 38.30 & 716.44 & 0.173m & 0.110m  \\
\midrule
Out-of-Domain & OpenDV-2K$^*$& 24.27& 442.67 & 0.238m & 0.136m\\
\bottomrule
\end{tabular}
\end{table}

\begin{table}[htbp]
\centering
 \caption{\textbf{Action instruction following under zero-shot evaluation.} $^*$  denotes the model is applied zero-shot to this dataset without fine-tuning.}
\label{tab:exp_aif}
\begin{tabular}{c|c|cc|cc}
\toprule
Training set & Test set& FID & FVD & $E^{\text{AIF}}_x$($\downarrow$) & $E^{\text{AIF}}_y$($\downarrow$) \\
\midrule
\methodname{} & OpenDV-2K$^*$& \textbf{24.27}& \textbf{442.67} & \textbf{0.238m} & \textbf{0.136m}\\
ONCE & OpenDV-2K$^*$& 28.37 & 473.59 & 0.255m & 0.23d9m\\
nuScenes & OpenDV-2K$^*$& 37.90 & 794.36 & 0.387m & 0.254m \\
\bottomrule
\end{tabular}
\end{table} 

\paragraph{Zero-shot evaluation.}
As shown in Table~\ref{tab:exp_aif}, we compared the performance of models trained on different datasets and their zero-shot generalization performance on new datasets. The results indicate that models trained on our dataset exhibit higher generation quality and significantly improved action-following ability. 
Especially, we noticed that richer driving actions in the autonomous driving datasets lead to significantly better AIF performance of models trained on them.
According to Figure \ref{fig:act_dist}, videos in \methodname{} averagely contain far richer driving actions compared to ONCE or nuScenes. 
This leads to the far better AIF performance of model trained on \methodname{} compared to those trained on ONCE or nuScenes.
we observed that the model trained on the ONCE dataset will always generate videos in which the vehicle moves in a straight line, even with action instructions to turn left/right or change lanes. This leads to its especially poor AIF performance in the lateral direction ($E^{\text{AIF}}_y$). We speculate that this is because the driving action of making turns or changing lanes is very rare in the ONCE dataset, as shown in Figure \ref{fig:act_dist}, which results in the lack of ability of the model trained on the ONCE dataset to follow the lateral motion instructions. Moreover, the even more lacking driving actions in the nuScenes dataset lead to a worse AIF performance of the world model.

\begin{figure}[htbp]
  \centering
  \includegraphics[width=1.0\textwidth]{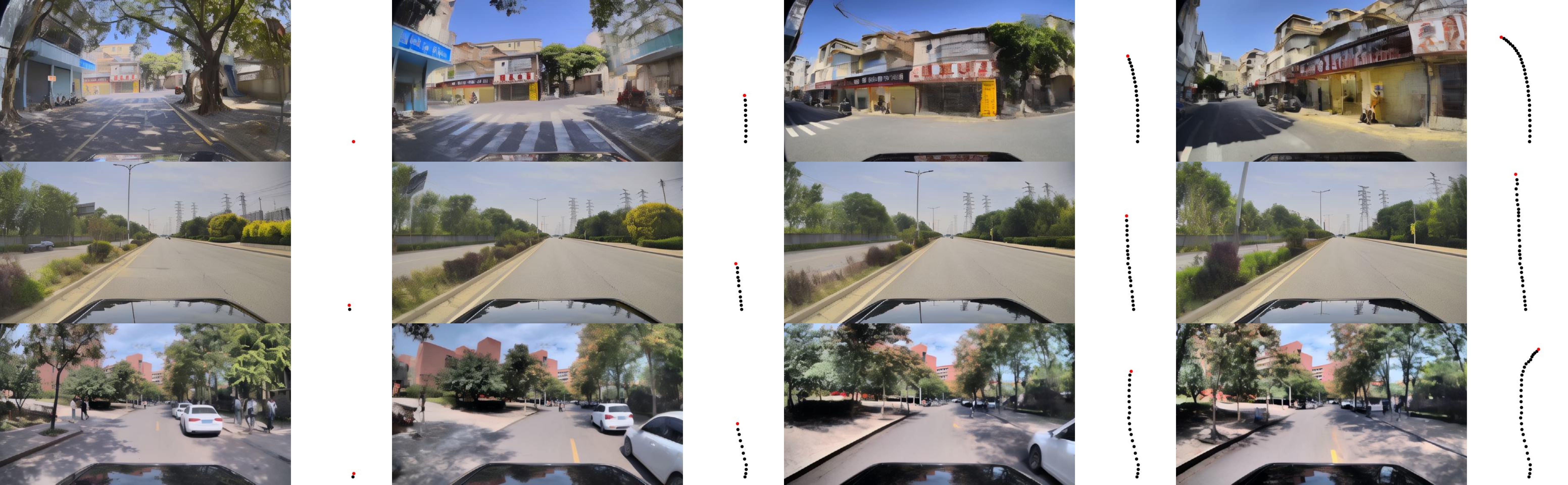}
  \caption{\textbf{Examples of ego trajectories estimated based on the generated videos.}}
  \label{fig:supp_aif}
\end{figure}

\paragraph{AIF visualization.}
We showcase examples of estimated trajectories from generated videos in Figure~\ref{fig:supp_aif}. In each frame, the red dot represents the current estimated camera pose and the black dots represent the camera poses in past frames. 

\subsection{Real-world Simulation}
\begin{figure}[htbp]
  \centering
\begin{subfigure}{0.565\linewidth}
    \includegraphics[width=\linewidth]{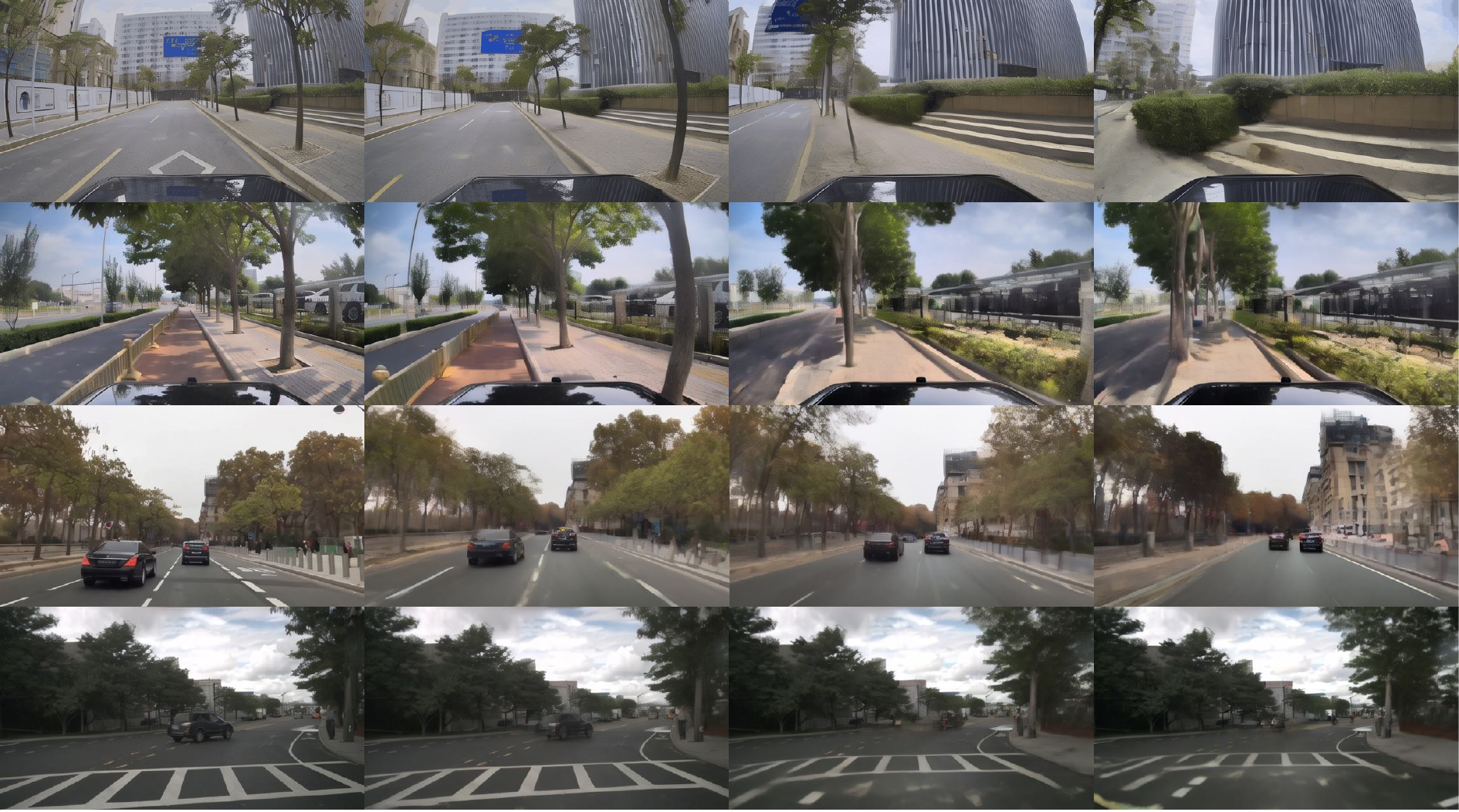}
    \caption{Action generalization}
     \label{fig:action_generalization}
    \end{subfigure}
    \begin{subfigure}{0.42\linewidth}
    \includegraphics[width=\linewidth]{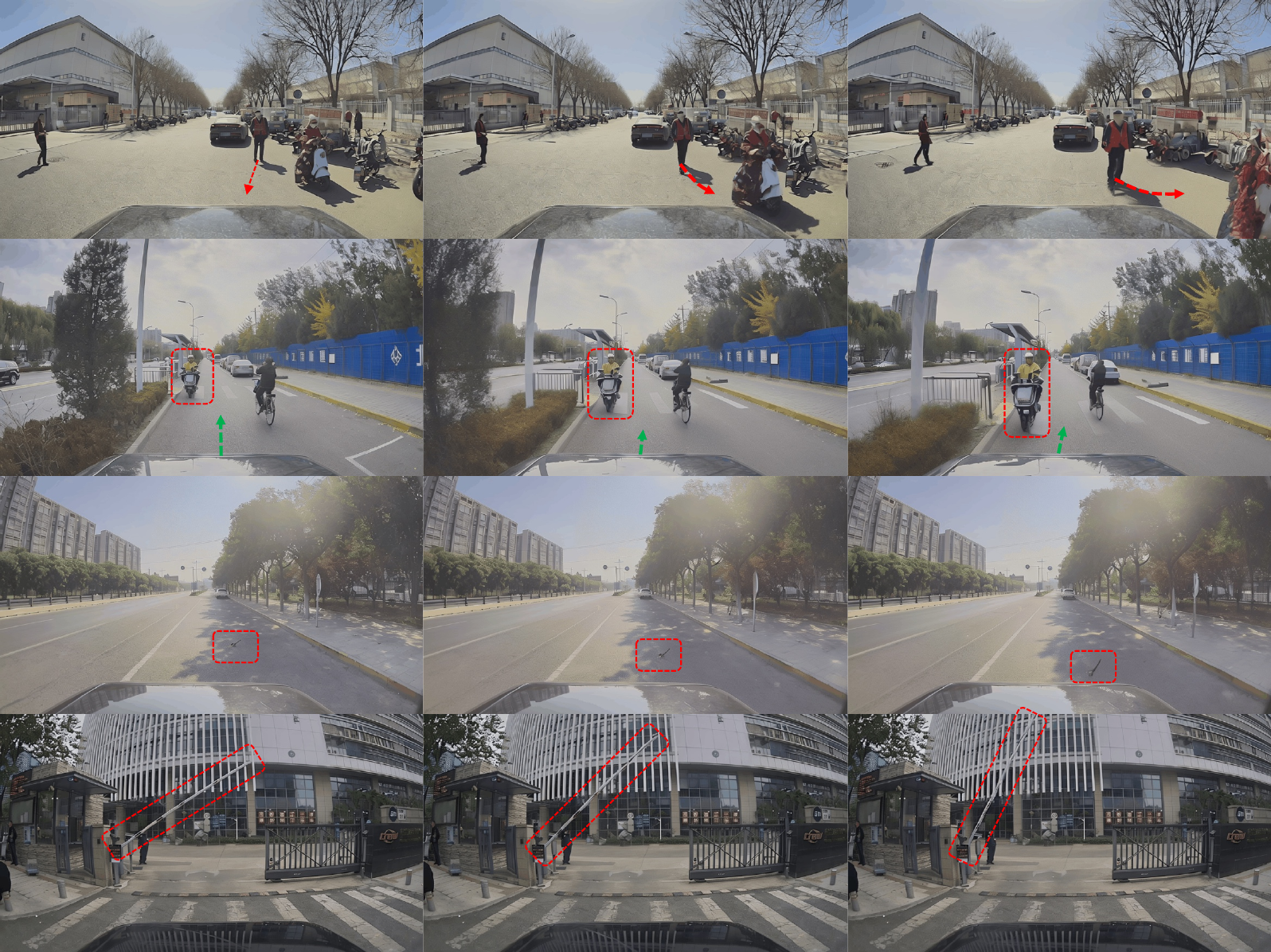}
    \caption{Interaction simulation}
    \label{fig:interaction_demo}
    \end{subfigure}
\caption{\textbf{Qualitative examples of our model's capability.}}
\vspace{-5mm}
\end{figure}

\noindent{\textbf{Action generalization.}}
Our model demonstrates robust generalization capabilities in two key aspects. As illustrated in Figure~\ref{fig:action_generalization}, firstly, it effectively generalizes to out-of-domain (OOD) actions, such as forcefully driving on pedestrian walkways, showcasing its adaptability to some unreasonable actions. Secondly, it successfully extends its capabilities to other datasets, executing tasks such as lane changes on the OpenDV-2K~\cite{yang2024generalized} dataset and backing-the-car maneuvers on the nuScenes~\cite{caesar2020nuscenes} dataset without requiring further fine-tuning. This underscores the model's potential as a real-world simulator, capable of adapting to diverse driving scenarios.

\noindent{\textbf{Dynamic agents.}}
We showcase our model's ability to simulate interactions with dynamic agents in Figure~\ref{fig:interaction_demo}. The results indicate that the model can provide reasonable responses based on our actions. The first scenario depicts a pedestrian opting to yield as our vehicle continues forward, resulting in a change in trajectory. In the second scenario, a delivery person opts to stop and wait at a narrow road.

\noindent{\textbf{Open-world dynamics.}}
In Figure~\ref{fig:interaction_demo}, our model showcases the simulations of rare scenarios encountered on the road, including interactions with moving birds and parking lot barriers.

\subsection{Limitations and Future Work}
\label{sec:limitation}
This dataset currently comprises only single-camera videos. Our primary focus is to maximize video diversity, which has led us to reduce the number of sensors used, enabling us to capture a wider range of scenes. Additionally, this paper primarily explores the value of the dataset, treating the model aspect as a baseline without any specialized design.
Although the \methodname{} dataset significantly improves model capabilities, there are still several limitations that require further investigation in future studies.

\begin{figure}[htbp]
  \centering
  \includegraphics[width=1.0\textwidth]{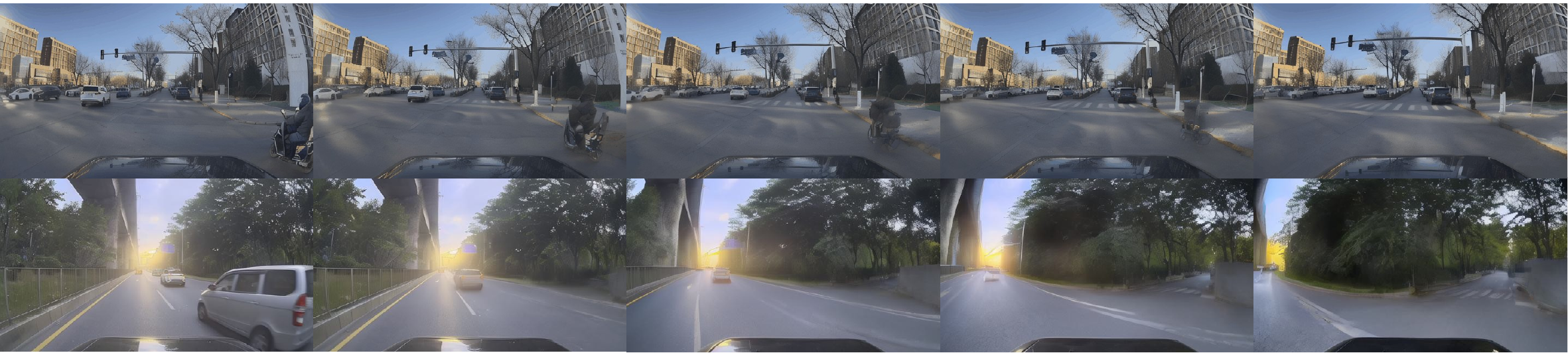} 
  \caption{\textbf{Examples of hallucination}. Top: object suddenly disappears. bottom: a non-existed road.}
  \label{fig:hallucination}
\end{figure}

\noindent{\textbf{Hallucination.}}
As shown in Figure~\ref{fig:hallucination}, we observed that the model exhibits some hallucinations, such as the sudden disappearance of objects, and when an action is unrealistic given the scene, such as forcefully turning right, the model sometimes imagines a new road.

\noindent{\textbf{Long-horizon visual prediction.}
Our baseline model is only capable of generating short videos, which can be used to simulate short-term interaction events. Longer predictions~\cite{brooks2022generating, yin2023nuwa, henschel2024streamingt2v} and faster generation~\cite{meng2023distillation,luo2023latent} are left for future research.

\noindent{\textbf{Driving policy.}}
The long-tail cases in our dataset are valuable for driving policy research. While this work focuses on visual prediction in world models, future studies can investigate how this data improves driving policy.

\section{Conclusion}
In this work, we present \methodname{}, a large-scale video dataset aimed at advancing the study of driving world models. \methodname{} offers a testbed for studying diverse real-world interactions. Our findings indicate that simulating interactions and rare dynamics observed in open-world environments remains an unsolved challenge, highlighting significant opportunities for future research.

\paragraph{Societal impacts.}
\label{sec:social}
By providing a comprehensive dataset covering diverse driving scenarios and behaviors, researchers can develop and refine algorithms that increase the safety, reliability, and efficiency of autonomous vehicles. However, the development of driving world model requires large and diverse driving videos, introducing privacy issues. 

\begin{ack}
We would like to thank Meituan's autonomous vehicle team for providing the data and computing resources. This work was supported in part by the National Key R\&D Program of China (No. 2022ZD0116500), the National Natural Science Foundation of China (No. U21B2042, No. 62320106010), and in part by the 2035 Innovation Program of CAS, and the InnoHK program, and in part by the Meituan Collaborative Research Project.
\end{ack}

\clearpage
\bibliographystyle{plain}
\bibliography{neurips_data_2024}
\clearpage

\section*{Checklist}

%%% BEGIN INSTRUCTIONS %%%
% The checklist follows the references.  Please
% read the checklist guidelines carefully for information on how to answer these
% questions.  For each question, change the default \answerTODO{} to \answerYes{},
% \answerNo{}, or \answerNA{}.  You are strongly encouraged to include a {\bf
% justification to your answer}, either by referencing the appropriate section of
% your paper or providing a brief inline description.  For example:
% \begin{itemize}
%   \item Did you include the license to the code and datasets? \answerYes{See Section.}
%   \item Did you include the license to the code and datasets? \answerNo{The code and the data are proprietary.}
%   \item Did you include the license to the code and datasets? \answerNA{}
% \end{itemize}
% Please do not modify the questions and only use the provided macros for your
% answers.  Note that the Checklist section does not count towards the page
% limit.  In your paper, please delete this instructions block and only keep the
% Checklist section heading above along with the questions/answers below.
%%% END INSTRUCTIONS %%%

\begin{enumerate}

\item For all authors...
\begin{enumerate}
  \item Do the main claims made in the abstract and introduction accurately reflect the paper's contributions and scope?
    \answerYes{See section~\ref{sec:intro}}
  \item Did you describe the limitations of your work?
    \answerYes{See section~\ref{sec:limitation}}
  \item Did you discuss any potential negative societal impacts of your work?
    \answerYes{See section~\ref{sec:social}}
  \item Have you read the ethics review guidelines and ensured that your paper conforms to them?
    \answerYes{}
\end{enumerate}

\item If you are including theoretical results...
\begin{enumerate}
  \item Did you state the full set of assumptions of all theoretical results?
    \answerNA{No theoretical results}
	\item Did you include complete proofs of all theoretical results?
    \answerNA{No theoretical results}
\end{enumerate}

\item If you ran experiments (e.g. for benchmarks)...
\begin{enumerate}
  \item Did you include the code, data, and instructions needed to reproduce the main experimental results (either in the supplemental material or as a URL)?
    \answerYes{In the supplemental material}
  \item Did you specify all the training details (e.g., data splits, hyperparameters, how they were chosen)?
    \answerYes{In the supplemental material}
	\item Did you report error bars (e.g., with respect to the random seed after running experiments multiple times)?
    \answerYes{We repeat evaluation multiple times and report the mean performance.}
	\item Did you include the total amount of compute and the type of resources used (e.g., type of GPUs, internal cluster, or cloud provider)?
    \answerYes{}
\end{enumerate}

\item If you are using existing assets (e.g., code, data, models) or curating/releasing new assets...
\begin{enumerate}
  \item If your work uses existing assets, did you cite the creators?
    \answerYes{}
  \item Did you mention the license of the assets?
    \answerYes{See supplemental material}
  \item Did you include any new assets either in the supplemental material or as a URL?
    \answerYes{See \url{https://drivingdojo.github.io}}
  \item Did you discuss whether and how consent was obtained from people whose data you're using/curating?
    \answerYes{The public release of the data has been approved and authorized by Meituan Inc.}
  \item Did you discuss whether the data you are using/curating contains personally identifiable information or offensive content?
    \answerYes{See Personal Identification Information Removal in Section~\ref{sec:data}}
\end{enumerate}

\item If you used crowdsourcing or conducted research with human subjects...
\begin{enumerate}
  \item Did you include the full text of instructions given to participants and screenshots, if applicable?
    \answerNA{}
  \item Did you describe any potential participant risks, with links to Institutional Review Board (IRB) approvals, if applicable?
    \answerNA{}
  \item Did you include the estimated hourly wage paid to participants and the total amount spent on participant compensation?
    \answerNA{}
\end{enumerate}

\end{enumerate}

%%%%%%%%%%%%%%%%%%%%%%%%%%%%%%%%%%%%%%%%%%%%%%%%%%%%%%%%%%%%

%%%%%%%%%%%%%%%%%%%%%%%%%%%%%%%%%%%%%%%%%%%%%%%%%%%%%%%%%%%%

\appendix
\clearpage
\noindent\textbf{\LARGE Appendix}
% \startcontents

% {
% \hypersetup{linkcolor=black}
% \printcontents{}{1}{}
% }
% \clearpage

\section{Dataset}
\subsection{Overview}
\label{sec:appendix_dataset}
We will publish the \methodname{} dataset, data format and annotation instructions, AIF benchmark, and code for the baseline method on our project page: \url{https://drivingdojo.github.io}.

\vspace{-3mm}
\paragraph{Terms of use and License.} Our dataset is released under the \textbf{CC BY-NC 4.0} license, allowing everyone to use it for non-commercial research purposes.

\vspace{-3mm}
\paragraph{Data maintenance.} The data is stored on Google Drive for global accessibility, and we will supply various links (e.g., Hugging Face) for researchers' convenience. We will maintain the data long-term and periodically verify its accessibility.

In the following, we showcase more video examples in our \methodname{} dataset, the corresponding videos are better illustrated on our project page. 

\begin{figure}[htbp]
  \centering
  \includegraphics[width=1.0\textwidth]{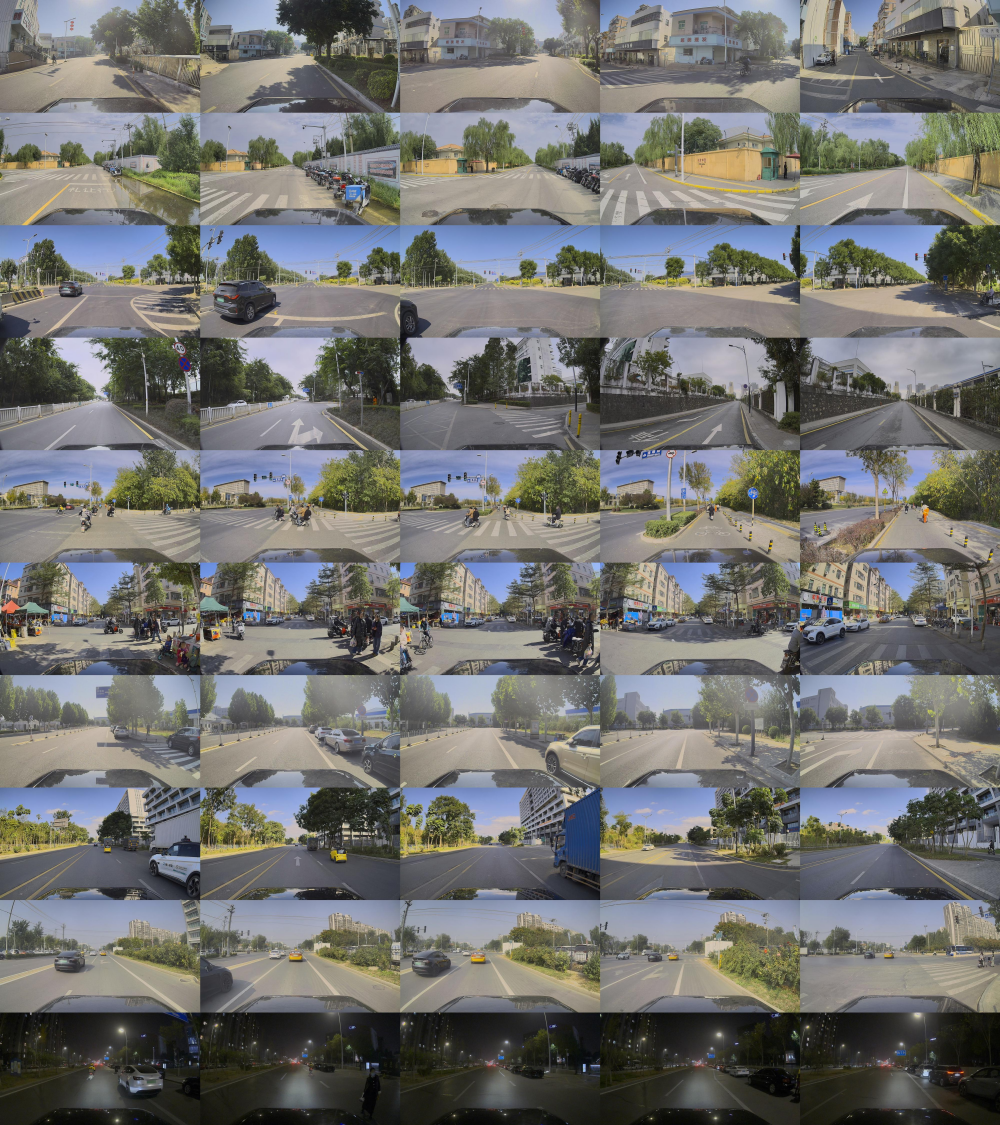}
  \caption{\textbf{Examples of rich ego-actions on the \methodname{} dataset.}}
  \label{fig:supp_action}
\end{figure}

\paragraph{Action completeness.}
We include more dataset visualizations depicting various ego-actions in Figure~\ref{fig:supp_action}. From top to bottom, the images show the ego vehicle performing left turns, right turns, going straight, lane-changing, and making emergency brakes during the driving.

\paragraph{Multi-agent interplay.}
\begin{figure}[htbp]
  \centering
  \includegraphics[width=1.0\textwidth]{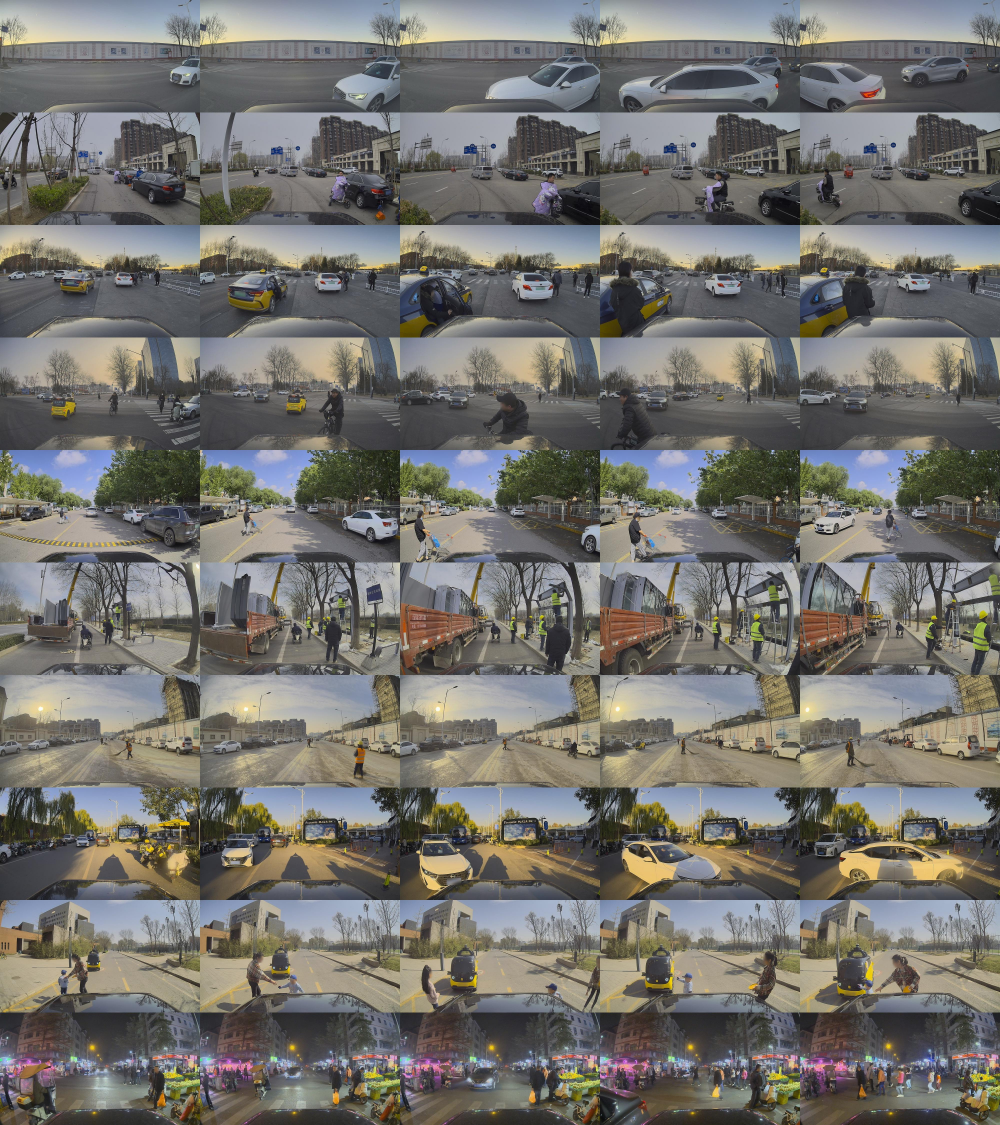}
  \caption{\textbf{Examples of multi-agent interplay on the \methodname{} dataset.}}
  \label{fig:supp_inter}
\end{figure}
Interaction plays a crucial role in driving scenarios. It usually means that the ego vehicle has engaged with other road users, leading to changes in the behavior of either the ego vehicle or the other road users.
As shown in Figure~\ref{fig:supp_inter}, we present a series of interaction examples in our dataset. In the first scenario, the car suddenly encounters another vehicle crossing its path while moving forward, prompting an abrupt braking maneuver.
The second scenario portrays the car encountering an electric scooter unexpectedly crossing its path. Illustrating the third scenario, the car comes across a vehicle in front opening its door, forcing an abrupt brake. In the fourth scenario, the challenge involves encountering a bicycle approaching from the opposite direction, while the fifth scenario involves navigating around a stroller. The sixth scenario showcases encountering road construction ahead, followed by encountering a street sweeper in the seventh scenario. The eighth scenario presents a situation where a car suddenly makes a U-turn from the opposite direction, prompting an urgent braking response from our vehicle. Subsequent scenarios involve interactions with pedestrians. 
These diverse interaction scenarios provide a crucial foundation for studying the interaction of real-world simulators.

\begin{figure}[htbp]
  \centering
  \includegraphics[width=1.0\textwidth]{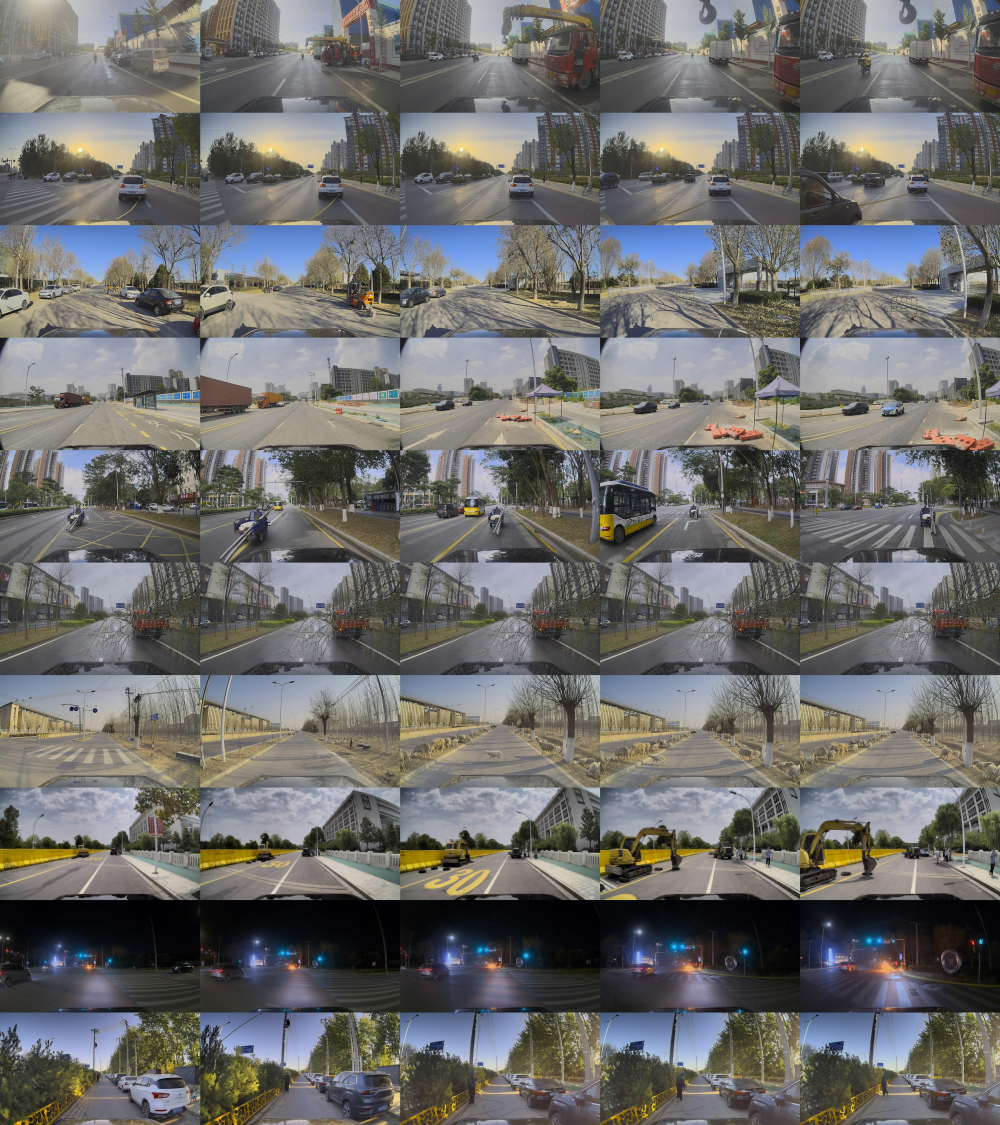}
  \caption{\textbf{Examples of diverse open-world objects on the \methodname{} dataset.}}
  \label{fig:supp_open}
\end{figure}
\paragraph{Open-world knowledge.}
In complex driving environments, we often encounter a wide variety of open-world situations. These scenarios can include sudden appearances of unexpected obstacles such as fallen trees, construction barriers, or abandoned vehicles. Typically belonging to the tail end of a long-tail distribution, these scenarios are rare yet crucial for ensuring safe driving.
In Figure~\ref{fig:supp_open}, we showcase a series of examples from the dataset, which fully demonstrate the richness of our dataset in capturing long-tail scenarios. From top to bottom, the examples illustrate encounters with a crane, a towing rope, construction barriers, a fallen roadblock, a vehicle transporting iron pipes, a vehicle transporting tree branches, a herd of sheep, an excavator, a bonfire, and power lines.

\subsection{Curation}
\label{sec:appendix_curation}
In this section, we provide the details of the curation procedure of each subset of \methodname{} dataset and the descriptions of curated actions and interactions. This section supplements the details for Section~\ref{sec:data} in the main paper.

\paragraph{Action Completeness}
Ego maneuvers for a car, particularly in the context of autonomous driving, refer to the actions and decisions the vehicle makes to navigate its environment safely and efficiently. 
Here is an exhaustive list of common ego maneuvers, and some examples in our datasets are shown in Figure~\ref{fig:actions} in the main paper:
\begin{itemize}[leftmargin=*]
    \item \textbf{Acceleration}:
     Increasing speed to match traffic flow.
    \item  \textbf{Deceleration}:
    Gradual slowing down for stop signs, traffic lights, or traffic congestion.
    \item \textbf{Lane Keeping}:
    Maintaining the current lane.
    \item  \textbf{Lane Changing}:
    Changing lanes to overtake slower vehicles or merge into traffic.
    \item \textbf{Turning}:
    Left/right or U-turns at intersections or roundabouts.
    \item \textbf{Stop and Move on}:
    Stopping/proceeding at traffic signals or stop signs. 
    \item \textbf{Emergency brake}: Abrupt and sudden braking maneuver to avoid a collision or mitigate the impact of a potential hazard.
\end{itemize}

So, in the \methodname{}-Action set, the videos follow different action commands, and the actions are mainly from the planning and control (PNC) signals, such as left and right turns, straight crossings, and lane changes. Each curated video clip begins with the PNC issuing a specific command and ends when the command is completed.

\paragraph{Multi-agent Interplay}
The examples of multi-agent interplay are shown in Figure~\ref{fig:interaction} in the main paper. Then we describe the detailed cases of the interactions with dynamic agents.
\begin{itemize}[leftmargin=*]
    \item \textbf{Cutting in/off}: Another vehicle abruptly changes lanes and enters the path of the autonomous vehicle. Ego vehicle changes lanes and enters the path of the other vehicles.
    \item \textbf{Meeting}: Ego vehicle encounters other vehicles traveling in the opposite direction.
    \item \textbf{Blocked}: Ego vehicle is stopped by other agents, such as vehicles, motorcycles, and pedestrians.
    \item \textbf{Overtaking and being overtaken}: Ego vehicle attempting to pass another vehicle and being passed by another vehicle. 
\end{itemize}

In the \methodname{}-Interplay set, the core data curation strategy is to find the interaction with other agents. The interaction is determined using PNC signals and manually defined rules.
The main interaction videos are from PNC dangerous interaction data. PNC conducts a deduction between the ego vehicle and obstacles. When the ego vehicle cannot avoid collision by turning the steering wheel or slowing down slightly, it is a PNC interaction case. 
\paragraph{Open-world Knowledge}

Here, we select some representative and interesting examples from these rare cases and show them in Figure \ref{fig:ood} in the main paper.
Based on the provided image and the given descriptions, here are the detailed descriptions of each rare case in autonomous driving:

(a) A worker's helmet rolls on the sidewalk next to the road. 
(b) A soccer ball is seen flying across the road. 
(c) A water bucket is depicted falling onto the road.
(d) Parcel boxes have fallen onto the road.
(e) A dog is crossing the road. 
(f) A rope is floating over the road. 
(g) The traffic light turns red.
(h) A boom barrier blocks the vehicle from moving forward.

As mentioned above, we curated \methodname{}-Open set in which the videos are more carefully categorized and labeled with text descriptions. The sources are unusual weather, foreign objects on the road surface, floating obstacles, falling objects, taking over cases, and interactions with traffic lights and boom barriers.  For curating the foreign objects/obstacles, we manually check and label them by a large number of data annotators.

\paragraph{Dataset Format}

DrivingDojo dataset provides a file named `dataset\_info.json' that stores information corresponding to each video segment, including the information shown in Table~\ref{tab:dataset_info}. The `type' represents the major category, `tag' represents the minor category, and `remark' provides detailed descriptions of the reasons for hard braking and intervention.

\begin{table}[htb]
\centering
\caption{The explanation of the information in dataset\_info.json.}
\label{tab:dataset_info}
\resizebox{\textwidth}{!}{
\begin{tabular}{c|c}
\toprule
\textbf{Information} & \textbf{Detailed explanation} \\
\midrule
meta\_info & weather, location, time, frame number \\ \midrule
description & type, tag, remark  \\ \midrule
videos & the image path for each frame \\  \midrule
camera\_info & the camera intrinsic parameters and extrinsic matrix for each frame \\ \midrule
action\_info & the coordinates of the next frame's camera position in the current camera coordinate system \\
\bottomrule
\end{tabular}}
\end{table}

The following is an example directory structure for a dataset:
\begin{verbatim}
.
dataset_info.json
action_info
    062959_s20-370_1712024694.0_1712024714.0
        0023_next_frame_position_at_current_camera.txt
        0025_next_frame_position_at_current_camera.txt
        0027_next_frame_position_at_current_camera.txt
        …
    145325_s20-190_1683790938.0_1683790958.0
        0024_next_frame_position_at_current_camera.txt
        0026_next_frame_position_at_current_camera.txt
        0028_next_frame_position_at_current_camera.txt
        …
    …
camera_info
    062959_s20-370_1712024694.0_1712024714.0
        0023_camera_parameters.txt
        0025_camera_parameters.txt
        0027_camera_parameters.txt
        …
    145325_s20-190_1683790938.0_1683790958.0
        0024_camera_parameters.txt
        0026_camera_parameters.txt
        0028_camera_parameters.txt
        …
    …
videos
    062959_s20-370_1712024694.0_1712024714.0
        0023_CameraFpgaP0H120.jpg
        0025_CameraFpgaP0H120.jpg
        0027_CameraFpgaP0H120.jpg
        …
    145325_s20-190_1683790938.0_1683790958.0
        0024_CameraFpgaP0H120.jpg
        0026_CameraFpgaP0H120.jpg
        0028_CameraFpgaP0H120.jpg
        …
    …
\end{verbatim}

\paragraph{Camera info.} The `camera info' refers to the extrinsic and intrinsic matrices of each frame of a fisheye camera. The world coordinate system is chosen as the East-North-Up (ENU) coordinate system. In the camera coordinate system, the x, y, and z axes respectively point to the right, down, and forward. We normalize the world coordinate system of the first frame to the origin, which means that the translation variables in the extrinsic matrices of each frame are subtracted by the translation variables of the first frame.

\paragraph{Action info.} The `action info' represents the coordinates of the next frame's camera position in the current camera coordinate system. Let the transformation matrix from the camera to the world coordinate system be $\left( \begin{array}{cc} R & T \\ 0^3 & 1 \end{array} \right)$. The calculation method for the action info $A_n$ of the $n$-th frame is shown in formula \ref{equation:action_info}. The orientation of xyz axes in matrix $A_n$ is consistent with the camera coordinate system, where the x, y, and z axes respectively point to the right, down, and forward.

\begin{equation}
\label{equation:action_info}
\left( \begin{array}{c}
A_n \\
1 \\
\end{array} \right)
=
\left( \begin{array}{cc}
R_n & T_n \\
0^3 & 1 \\
\end{array} \right)^{-1}
\left( \begin{array}{c}
T_{n+1} \\
1 \\
\end{array} \right)
\end{equation}

\begin{figure}[htbp]
  \centering
  \includegraphics[width=0.95\textwidth]{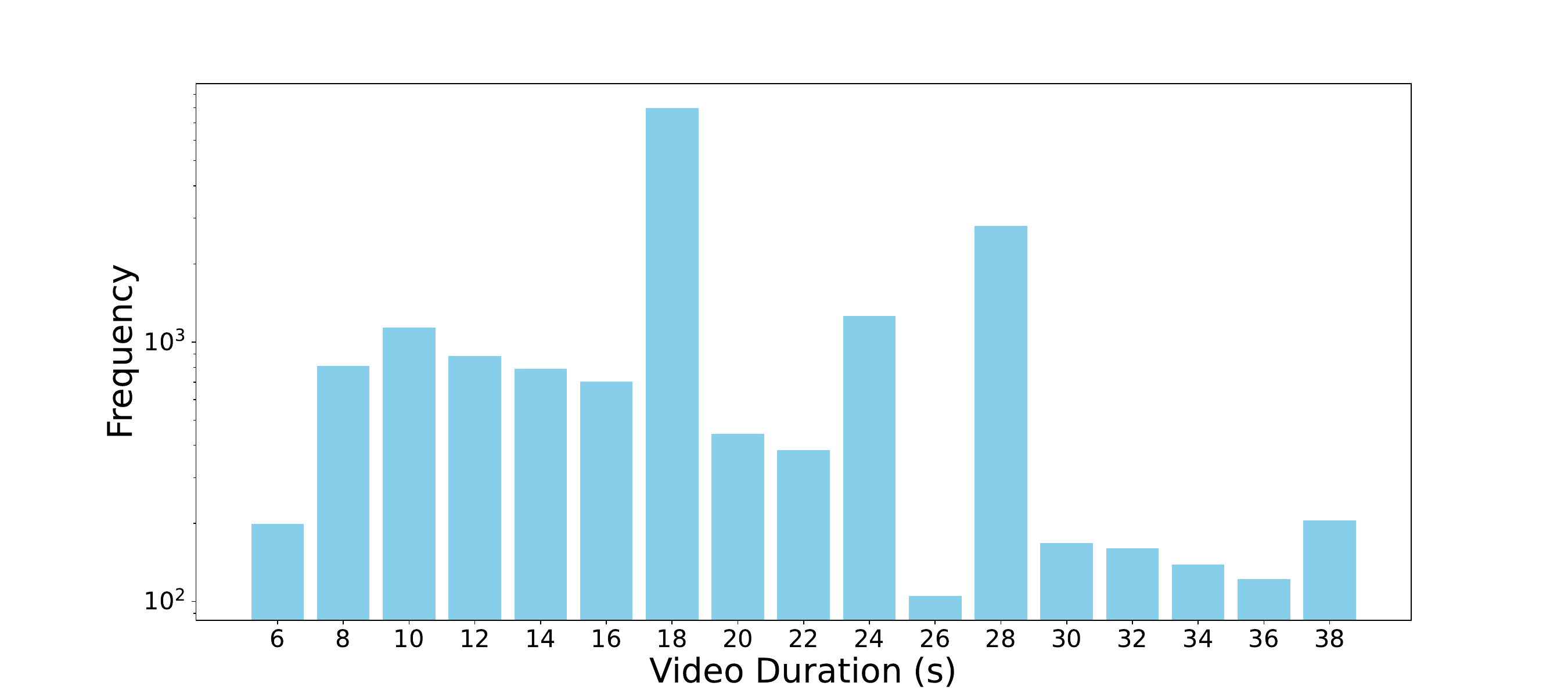}
  \caption{\textbf{Distribution of \methodname{} video duration.}}
  \label{fig:duration}
\end{figure}
\paragraph{Video info.} 
The video is stored as a sequence of individual image frames. The distribution of video duration is shown in Figure~\ref{fig:duration}, with the majority of videos lasting around 20 seconds.

% --------------------------------------------------------------
\section{Implementation Details}

\subsection{Experiment Setup}
During the experiment, we employed two settings for training the model.
In the first setting, the focus is on visual prediction: the model predicts subsequent video content based solely on the initial frame image. In the second setting, we employ action-controlled video generation. Here, alongside the initial frame image, action information for the subsequent frames is provided to the model, enabling it to predict the ensuing video content.

\paragraph{Visual prediction.} 
In this setup, we trained a high-resolution version of the model, 1024$\times$576 resolution for 14 frames, aimed at better capturing the generation of long-tail objects. Additionally, we developed a low-resolution version of the model, 576$\times$320 resolution for 30 frames, to simulate various vehicle behaviors and interaction events. We fine-tune all parameters of the U-Net model.

\paragraph{Action instruction following.} 
In this setup, we trained model using 576$\times$320 resolution for 30 frames. We fine-tune all U-Net parameters together with a new action encoder.

\subsection{Training}
\label{sec:appendix_training}
We initialize the model using the SVD-XT checkpoint. Following SVD, our model is trained with the EDM framework~\cite{karras2022elucidating}. 
During training, we set the $\text{fps}$ to 5 and the $\text{motion\_bucket\_id}$ to 127. We utilize the AdamW optimizer~\cite{loshchilov2018decoupled} with a learning rate of $1\times 10^{-5}$. The training process is conducted on 16 NVIDIA A100 (80G) GPUs with 32 batch size for 50K iterations. To allow classifier-free guidance~\cite{ho2021classifier}, we drop out action feature with a ratio of 20\%.

\subsection{Evaluation}
During inference, we generate videos using the DDIM sampler for 25 steps.

\paragraph{Visual Quality.}
To evaluate the quality of the generated video, we utilize FID (Frechet Inception Distance)~\cite{heusel2017gans} and FVD (Frechet Video Distance)~\cite{unterthiner2018towards} as the main metrics. For FID calculation on videos, we randomly select 5,000 frames for evaluation. Additionally, for FVD calculation, we generate 256 videos for evaluation. The results are the average of 10 calculations. We use the official UCF FVD evaluation code\footnote{\url{https://github.com/SongweiGe/TATS/}}. 

\paragraph{Action instruction following (AIF).}
For each generated video with action instructions, we estimate the camera poses for each frame in the video, align the scale of the estimated trajectory with the instruction trajectory, and compare the vehicle motion in each frame with the respective action instructions.
We estimate the ego trajectories in generated videos using the offline visual structure-from-motion (SfM) implementation COLMAP~\cite{schoenberger2016sfm,schoenberger2016mvs}. We found that moving objects significantly impact the quality of the reconstruction, so we used instance masks to occlude foreground moving objects during the reconstruction process.
For videos generated based on initial images from \methodname{}, we fix the camera intrinsic parameters as the ground truth values for videos from \methodname{}. For videos generated from initial images with unknown camera intrinsics (e.g. images from OpenDV-2K), we estimate the camera intrinsics together with the camera extrinsics of images. 
We perform feature point extraction, feature point matching, and sparse scene reconstruction with the official implementation of COLMAP\footnote{\url{https://github.com/colmap/colmap}} to estimate the poses of cameras.
In our experiments, we generate videos in 30 frames and align the scale of estimated trajectories with the instruction trajectories based on the motions in the first $N=10$ frames.
We report the mean value of the absolute error between estimated motions and instruction motions in all video frames. 

\section{Visualizations}
In this section, we show the model generation demos trained on the \methodname{} dataset. As shown in Figure~\ref{fig:supp_demo_image}, our model can generate high-resolution, complex driving scenarios.

\begin{figure}[htbp]
  \centering
  \includegraphics[width=1.0\textwidth]{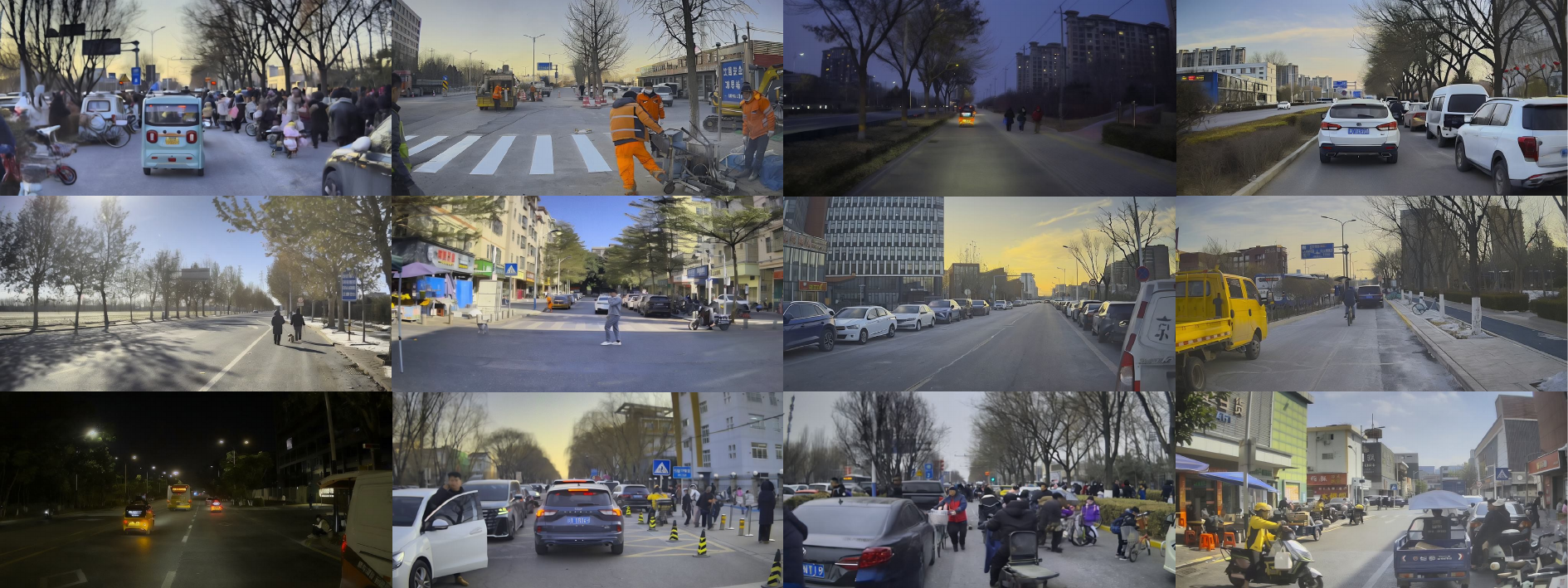}
  \caption{\textbf{Examples of high-resolution and complex scenarios generation.} For illustration purposes, we represent each video example with a single frame.}
  \label{fig:supp_demo_image}
\end{figure}

\begin{figure}[htbp]
  \centering
  \includegraphics[width=1.0\textwidth]{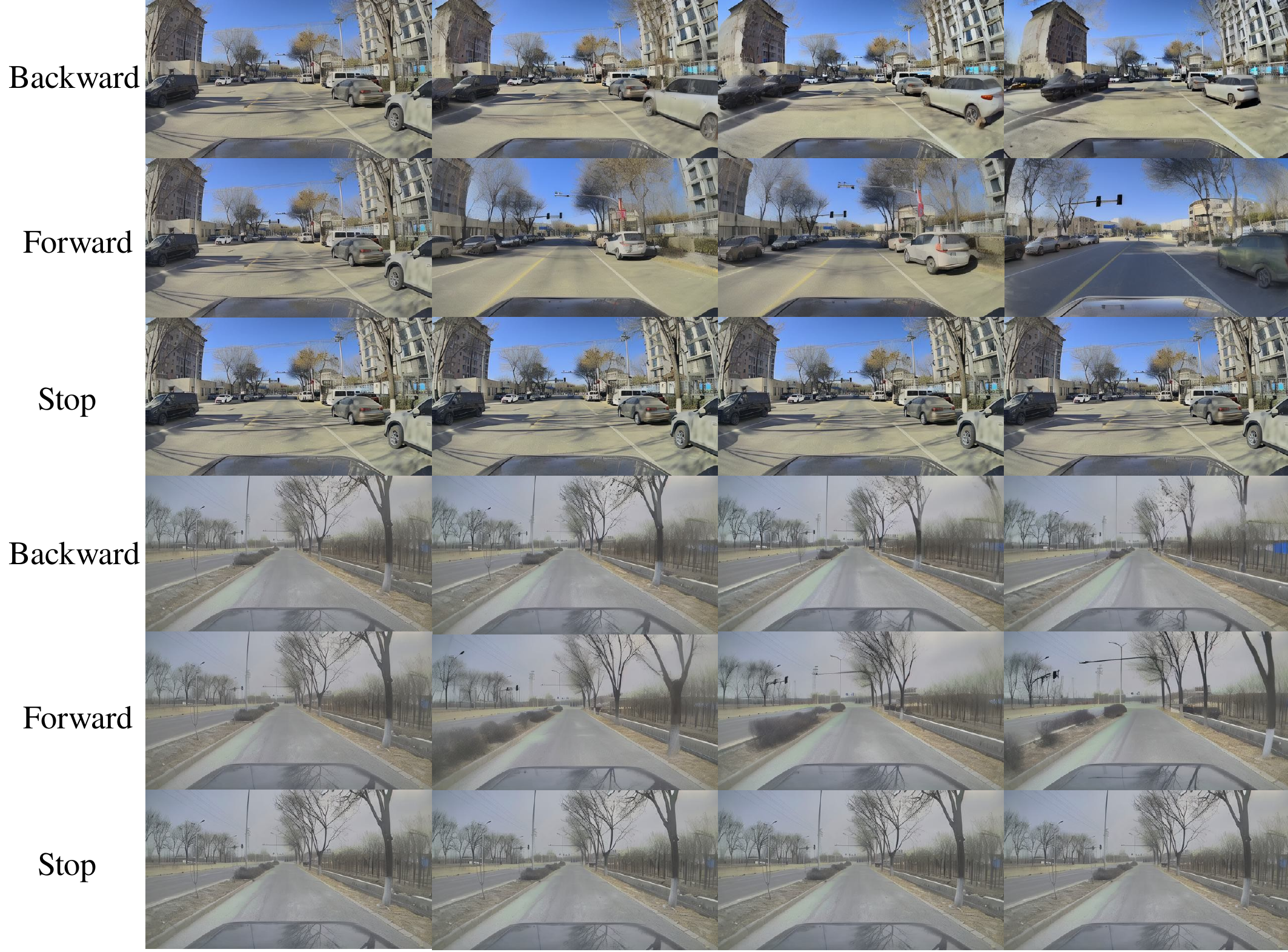}
  \caption{\textbf{Examples of diverse action-based video generation.}}
  \label{fig:supp_demo_action1}
\end{figure}

\subsection{Diverse Actions}
As shown in Figure~\ref{fig:supp_demo_action1}, we demonstrate how actions control the generation of different futures, such as moving forward, backward, and stopping.

\subsection{Dynamic Interaction}
As shown in Figure~\ref{fig:supp_demo_inter}, we observe that choosing different actions can influence the behavior of other vehicles, resulting in different responses from the world model. For instance, in the first example, if we choose to proceed slowly, the vehicle on the left decides to stop and yield. Conversely, if our vehicle stops, the left vehicle perceives an obstruction and slightly reverses to make way. In the second example, when we choose to brake, the right vehicle quickly cuts in front of us, while if we choose to proceed straight, the right vehicle waits in place.
\begin{figure}[htbp]
  \centering
  \includegraphics[width=1.0\textwidth]{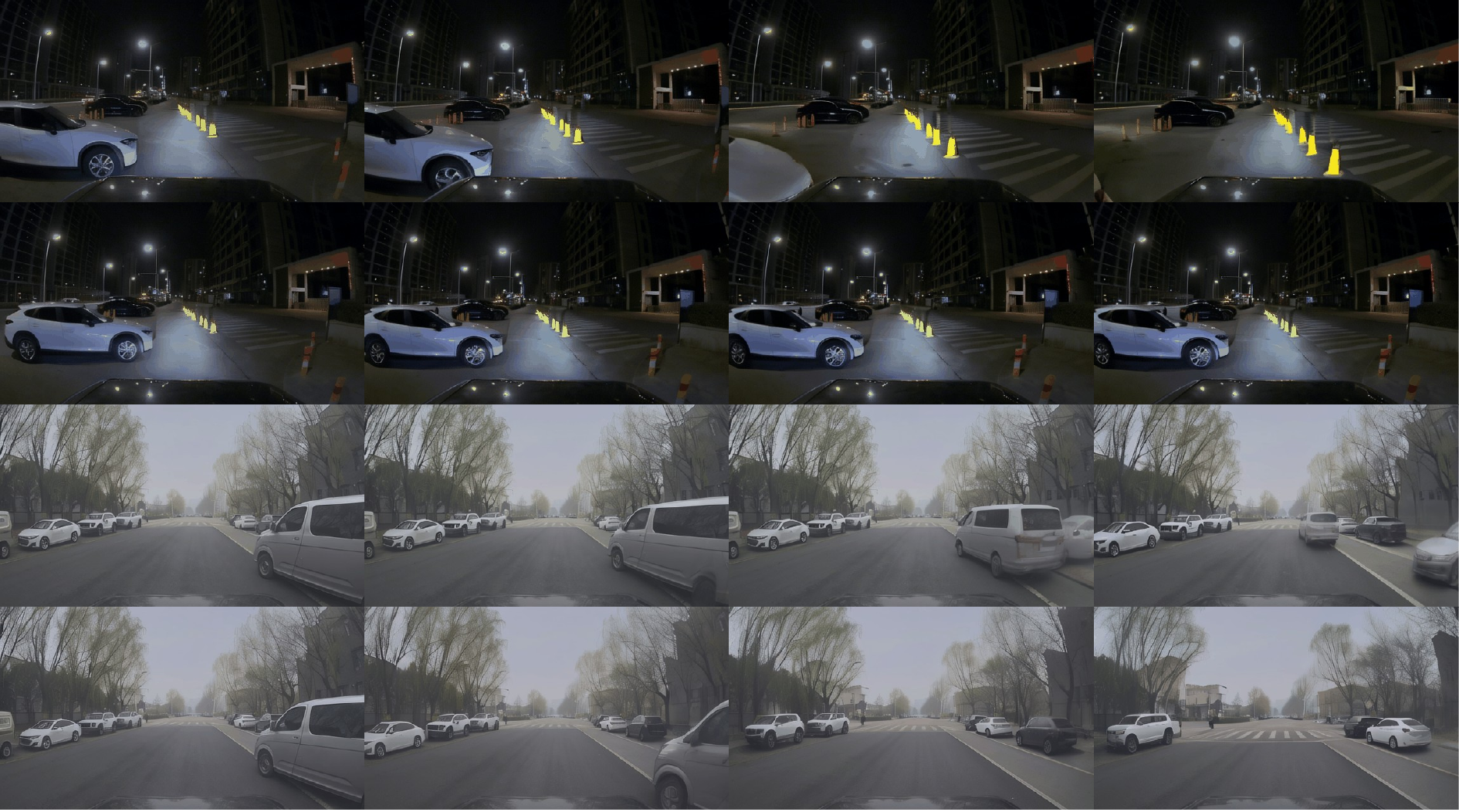}
  \caption{\textbf{Simulation of interaction with other agents.}}
  \label{fig:supp_demo_inter}
\end{figure}

\subsection{Open-world Knowledge}
As illustrated in Figure~\ref{fig:supp_demo_open}, we demonstrate the model's ability to simulate various open-world objects, such as encountering construction zones, rare objects like ladders or balloons on the road, and simulating a puddle of water on the ground.
\begin{figure}[htbp]
  \centering
  \includegraphics[width=1.0\textwidth]{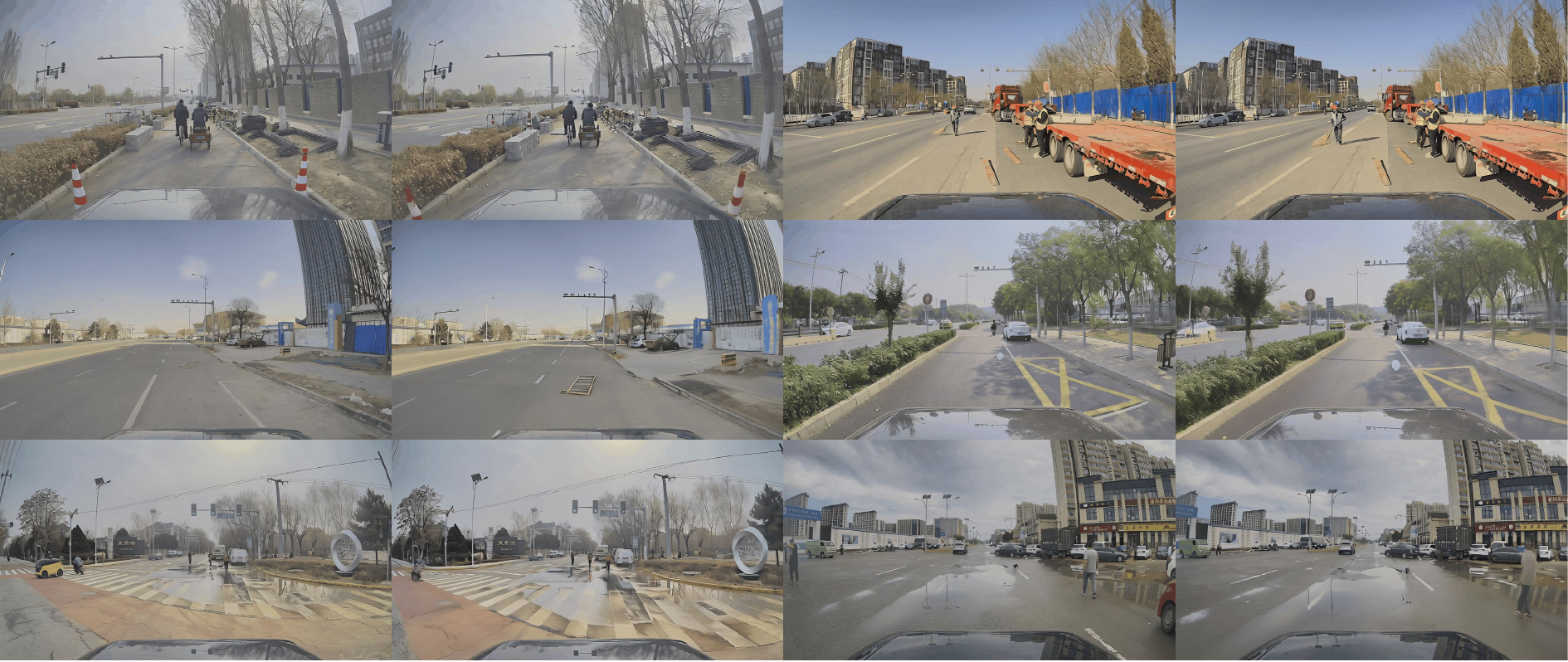}
  \caption{\textbf{Simulation of various open-world objects on the road.}}
  \label{fig:supp_demo_open}
\end{figure}

% ---------------------------------------------------------------
\section{License of Assets}
We report licenses of all artifacts used in this work in this section.

\noindent\textbf{Model}
We use the pre-trained stable video diffusion~\cite{blattmann2023stable} checkpoints from the huggingface platform. 
These checkpoints are released under the stable video diffusion non-commercial community license agreement\footnote{\url{https://huggingface.co/stabilityai/stable-video-diffusion-img2vid-xt/blob/main/LICENSE}} for research purpose.

\noindent\textbf{Our Dataset}
Our dataset is collected and curated by the autonomous driving team of Meituan Inc.
The road test and data collection procedures conform to privacy and security requirements of local authorities.
The authors have obtained the permission for publicly releasing this dataset from both the management team and the company legal team.
All personal identifiable information has been removed by both algorithm and subsequent manual inspection.
We release the dataset under the CC BY-NC 4.0 license.

\noindent\textbf{Other Datasets}
We use other public datasets in this work including nuScenes~\cite{caesar2020nuscenes}, ONCE~\cite{mao2021one} and OpenDV-2k~\cite{yang2024generalized}.
The nuScenes~\cite{caesar2020nuscenes} dataset is released under the CC BY-NC-SA 4.0 license with Dataset Terms~\footnote{\url{https://www.nuscenes.org/terms-of-use}}.
The ONCE dataset is also released under the CC BY-NC-SA 4.0 license with Dataset Terms~\footnote{\url{https://once-for-auto-driving.github.io/terms_of_use.html}}.
The OpenDV-2K dataset is constructed from publicly licensed datasets and youtube videos that the authors claimed to support academic usage licenses. 

% ---------------------------------------------------------------
% \clearpage

\section{Datasheet}
\subsection{Motivation}

\begin{itemize}

\item \textbf{For what purpose was the dataset created?} Was there a specific task in mind? Was there a specific gap that needed to be filled? Please provide a description.

We introduce \methodname{}, the first dataset tailor-made for training interactive world models with complex driving dynamics. 
Our dataset features video clips with a complete set of driving maneuvers, diverse multi-agent interplay, and rich open-world driving knowledge, laying a stepping stone for future world model development.

\item \textbf{Who created the dataset (e.g., which team, research group) and on behalf of which entity (e.g., company, institution, organization)?}

Institute of Automation, Chinese Academy of Sciences, University of Chinese Academy of Sciences, Meituan Inc., and Centre for Artificial Intelligence and Robotics, HKISI\_CAS.

\item \textbf{Who funded the creation of the dataset?}If there is an associated grant, please provide the name of the grantor and the grant name and number.

This work was supported in part by the National Key R\&D Program of China (No. 2022ZD0116500), the National Natural Science Foundation of China (No. U21B2042, No. 62320106010), and in part by the 2035 Innovation Program of CAS, and the InnoHK program.

\item \textbf{Any other comments?}

No.

\end{itemize}

\subsection{Composition}

% Dataset creators should read through these questions prior to
% any data collection and then provide answers once data collection is
% complete. Most of the questions in this section are intended to
% provide dataset consumers with the information they need to make
% informed decisions about using the dataset for their chosen
% tasks. Some of the questions are {designed to elicit} information
% about compliance with the EU's General Data Protection Regulation
% (GDPR) or comparable regulations in other jurisdictions.

% {Questions that apply only to datasets that relate to people are
% grouped together at the end of the section. We recommend taking a
% broad interpretation of whether a dataset relates to people. For
% example, any dataset containing text that was written by people
% relates to people.}\\

\begin{itemize}

\item \textbf{What do the instances that comprise the dataset
    represent (e.g., documents, photos, people, countries)?} Are there
  multiple types of instances (e.g., movies, users, and ratings;
  people and interactions between them; nodes and edges)? Please
  provide a description.

The instances of our \methodname{} dataset are videos with ego actions and \methodname{}-Open subset is also with text descriptions for each scene.
\item \textbf{How many instances are there in total (of each type, if appropriate)?}

There are 17.8k videos for the whole \methodname{} dataset, in which the \methodname{}-Action subset has 7.9k videos, \methodname{}-Interplay subset has 6.2k videos, and \methodname{}-Open has 3.7k videos.
\item \textbf{Does the dataset contain all possible instances or is it
    a sample (not necessarily random) of instances from a larger set?}
  If the dataset is a sample, then what is the larger set? Is the
  sample representative of the larger set (e.g., geographic coverage)?
  If so, please describe how this representativeness was
  validated/verified. If it is not representative of the larger set,
  please describe why not (e.g., to cover a more diverse range of
  instances, because instances were withheld or unavailable).

The \methodname{} dataset is sampled from a data pool of around 7500 hours. About representativeness, please refer to the Data Curation section (Sec.~\ref{sec:data} and Sec.~\ref{sec:appendix_curation}).
\item \textbf{What data does each instance consist of?} ``Raw'' data
  (e.g., unprocessed text or images) or features? In either case,
  please provide a description.

 \methodname{}-Action and \methodname{}-Interplay subsets consist of videos and ego actions, and \methodname{}-Open subset consists of videos, ego actions, and text descriptions.
\item \textbf{Is there a label or target associated with each
    instance?} If so, please provide a description.

Yes. There is a text description label for each instance in \methodname{}-Open subset, which describes the open-world knowledge in the scene.
\item \textbf{Is any information missing from individual instances?}
  If so, please provide a description, explaining why this information
  is missing (e.g., because it was unavailable). This does not include
  intentionally removed information, but might include, e.g., redacted
  text.

No.
\item \textbf{Are relationships between individual instances made
    explicit (e.g., users' movie ratings, social network links)?} If
  so, please describe how these relationships are made explicit.

No.
\item \textbf{Are there recommended data splits (e.g., training,
    development/validation, testing)?} If so, please provide a
  description of these splits, explaining the rationale behind them.

No. There is no need for the validation/testing split. We care about zero-shot generation.
\item \textbf{Are there any errors, sources of noise, or redundancies
    in the dataset?} If so, please provide a description.

Yes. The sources of noise may be inaccurate poses, camera noises, and human-sourced text noises.
\item \textbf{Is the dataset self-contained, or does it link to or
    otherwise rely on external resources (e.g., websites, tweets,
    other datasets)?} If it links to or relies on external resources,
    a) are there guarantees that they will exist, and remain constant,
    over time; b) are there official archival versions of the complete
    dataset (i.e., including the external resources as they existed at
    the time the dataset was created); c) are there any restrictions
    (e.g., licenses, fees) associated with any of the external
    resources that might apply to a {dataset consumer}? Please provide
    descriptions of all external resources and any restrictions
    associated with them, as well as links or other access points, as
    appropriate.

Yes. the \methodname{} dataset is self-contained.
\item \textbf{Does the dataset contain data that might be considered
    confidential (e.g., data that is protected by legal privilege or
    by doctor{--}patient confidentiality, data that includes the content
    of individuals' non-public communications)?} If so, please provide
    a description.

No.
\item \textbf{Does the dataset contain data that, if viewed directly,
    might be offensive, insulting, threatening, or might otherwise
    cause anxiety?} If so, please describe why.

No.
\end{itemize}

% {If the dataset does not }relate to people, you may skip the remaining questions in this section.

% \begin{itemize}

% \item \textbf{Does the dataset identify any subpopulations (e.g., by
%     age, gender)?} If so, please describe how these subpopulations are
%   identified and provide a description of their respective
%   distributions within the dataset.

% \item \textbf{Is it possible to identify individuals (i.e., one or
%     more natural persons), either directly or indirectly (i.e., in
%     combination with other data) from the dataset?} If so, please
%     describe how.

% \item \textbf{Does the dataset contain data that might be considered
%     sensitive in any way (e.g., data that reveals rac{e} or ethnic
%     origins, sexual orientations, religious beliefs, political
%     opinions or union memberships, or locations; financial or health
%     data; biometric or genetic data; forms of government
%     identification, such as social security numbers; criminal
%     history)?} If so, please provide a description.

% \item \textbf{Any other comments?}

% \end{itemize}

\subsection{Collection Process}

% As with the {questions in the} previous section, dataset creators should
% read through these questions prior to any data collection to flag
% potential issues and then provide answers once collection is complete.
% {In addition to the goals outlined in the previous section, the
% questions in this section are designed to elicit information that may
% help researchers and practitioners to create alternative datasets with
% similar characteristics. Again, questions that apply only to datasets
% that relate to people are grouped together at the end of the
% section.}\\

\begin{itemize}

\item \textbf{How was the data associated with each instance
    acquired?} Was the data directly observable (e.g., raw text, movie
  ratings), reported by subjects (e.g., survey responses), or
  indirectly inferred/derived from other data (e.g., part-of-speech
  tags, model-based guesses for age or language)? If the data was reported
  by subjects or indirectly inferred/derived from other data, was the
  data validated/verified? If so, please describe how.

\methodname{} dataset is collected using the platform of Meituan's autonomous delivery vehicles. 

\item \textbf{What mechanisms or procedures were used to collect the
    data (e.g., hardware apparatuses or sensors, manual human
    curation, software programs, software APIs)?} How were these
    mechanisms or procedures validated?

\methodname{} dataset is collected using the platform of Meituan's autonomous delivery vehicles with fish-eye RGB cameras. The cameras are calibrated. The text labels are manually validated.
\item \textbf{If the dataset is a sample from a larger set, what was
    the sampling strategy (e.g., deterministic, probabilistic with
    specific sampling probabilities)?}

Please refer to the Data Curation section (Sec.~\ref{sec:data} and Sec.~\ref{sec:appendix_curation}).
\item \textbf{Who was involved in the data collection process (e.g.,
    students, crowdworkers, contractors) and how were they compensated
    (e.g., how much were crowdworkers paid)?}

The data collectors are employed by Meituan Inc. and are paid by Meituan Inc.
\item \textbf{Over what timeframe was the data collected?} Does this
  timeframe match the creation timeframe of the data associated with
  the instances (e.g., recent crawl of old news articles)?  If not,
  please describe the timeframe in which the data associated with the
  instances was created.

The data are collected from May 2022 to May 2024. This
  timeframe matches the creation timeframe of the data associated with the instances.
\item \textbf{Were any ethical review processes conducted (e.g., by an
    institutional review board)?} If so, please provide a description
  of these review processes, including the outcomes, as well as a link
  or other access point to any supporting documentation.
  
Yes. The ethical review is conducted before the release by Meituan Inc.
\end{itemize}

% {If the dataset does not relate to people, you may skip the remaining questions in this section.}

% \begin{itemize}

% \item \textbf{Did you collect the data from the individuals in
%     question directly, or obtain it via third parties or other sources
%     (e.g., websites)?}

% \item \textbf{Were the individuals in question notified about the data
%     collection?} If so, please describe (or show with screenshots or
%   other information) how notice was provided, and provide a link or
%   other access point to, or otherwise reproduce, the exact language of
%   the notification itself.

% \item \textbf{Did the individuals in question consent to the
%     collection and use of their data?} If so, please describe (or show
%   with screenshots or other information) how consent was requested and
%   provided, and provide a link or other access point to, or otherwise
%   reproduce, the exact language to which the individuals consented.

% \item \textbf{If consent was obtained, were the consenting individuals
%     provided with a mechanism to revoke their consent in the future or
%     for certain uses?} If so, please provide a description, as well as
%   a link or other access point to the mechanism (if appropriate).

% \item \textbf{Has an analysis of the potential impact of the dataset
%     and its use on data subjects (e.g., a data protection impact
%     analysis) been conducted?} If so, please provide a description of
%   this analysis, including the outcomes, as well as a link or other
%   access point to any supporting documentation.

% \item \textbf{Any other comments?}

% \end{itemize}

\subsection{Preprocessing/cleaning/labeling}

% Dataset creators should read through these questions prior to any
% preprocessing, cleaning, or labeling and then provide answers once
% these tasks are complete. The questions in this section are intended
% to provide dataset consumers with the information they need to
% determine whether the ``raw'' data has been processed in ways that are
% compatible with their chosen tasks. For example, text that has been
% converted into a ``bag-of-words'' is not suitable for tasks involving
% word order.\\

\begin{itemize}

\item \textbf{Was any preprocessing/cleaning/labeling of the data done
    (e.g., discretization or bucketing, tokenization, part-of-speech
    tagging, SIFT feature extraction, removal of instances, processing
    of missing values)?} If so, please provide a description. If not,
  you may skip the remain{ing} questions in this section.

No. 
\item \textbf{Was the ``raw'' data saved in addition to the preprocessed/cleaned/labeled data (e.g., to support unanticipated future uses)?} If so, please provide a link or other access point to the ``raw'' data.

N/A.
\item \textbf{Is the software {that was} used to preprocess/clean/label the {data} available?} If so, please provide a link or other access point.

N/A.
\item \textbf{Any other comments?}

No.
\end{itemize}

\subsection{Uses}

% {The} questions {in this section} are intended to encourage dataset
% creators to reflect on the tasks for which the dataset should and
% should not be used. By explicitly highlighting these tasks, dataset
% creators can help dataset consumers to make informed decisions,
% thereby avoiding potential risks or harms.\\

\begin{itemize}

\item \textbf{Has the dataset been used for any tasks already?} If so, please provide a description.

The \methodname{} dataset has been used for driving world models. The experiments are in Sec. 5 in the main paper and Sec. C in the appendix.
\item \textbf{Is there a repository that links to any or all papers or systems that use the dataset?} If so, please provide a link or other access point.

Yes. Please refer to the webset:\url{https://drivingdojo.github.io}.
\item \textbf{What (other) tasks could the dataset be used for?}

The \methodname{} dataset could be used for training end-to-end autonomous driving models.
\item \textbf{Is there anything about the composition of the dataset or the way it was collected and preprocessed/cleaned/labeled that might impact future uses?} For example, is there anything that a {dataset consumer} might need to know to avoid uses that could result in unfair treatment of individuals or groups (e.g., stereotyping, quality of service issues) or other {risks or} harms (e.g., {legal risks,} financial harms{)?} If so, please provide a description. Is there anything a {dataset consumer} could do to mitigate these {risks or} harms?

No.
\item \textbf{Are there tasks for which the dataset should not be used?} If so, please provide a description.

Due to the known biases of the dataset, under no circumstance should any models be put into production using the dataset as is. It is neither safe nor responsible. As it stands, the dataset should be solely used for research purposes in its uncurated state.
\item \textbf{Any other comments?}

No.
\end{itemize}

\subsection{Distribution}

% Dataset creators should provide answers to these questions prior to
% distributing the dataset either internally within the entity on behalf
% of which the dataset was created or externally to third parties.\\

\begin{itemize}

\item \textbf{Will the dataset be distributed to third parties outside of the entity (e.g., company, institution, organization) on behalf of which the dataset was created?} If so, please provide a description.

Yes, the dataset will be open-source.
\item \textbf{How will the dataset will be distributed (e.g., tarball on website, API, GitHub)?} Does the dataset have a digital object identifier (DOI)?

On our website: \url{https://drivingdojo.github.io}.
\item \textbf{When will the dataset be distributed?}

We have released some demos on the project page. The whole \methodname{} dataset will be public in the camera-ready version.
\item \textbf{Will the dataset be distributed under a copyright or other intellectual property (IP) license, and/or under applicable terms of use (ToU)?} If so, please describe this license and/or ToU, and provide a link or other access point to, or otherwise reproduce, any relevant licensing terms or ToU, as well as any fees associated with these restrictions.

\methodname{} dataset will be distributed under the CC BY-NC 4.0 license.
\item \textbf{Have any third parties imposed IP-based or other restrictions on the data associated with the instances?} If so, please describe these restrictions, and provide a link or other access point to, or otherwise reproduce, any relevant licensing terms, as well as any fees associated with these restrictions.

No.
\item \textbf{Do any export controls or other regulatory restrictions apply to the dataset or to individual instances?} If so, please describe these restrictions, and provide a link or other access point to, or otherwise reproduce, any supporting documentation.

No.
\item \textbf{Any other comments?}

No.
\end{itemize}

\subsection{Maintenance}

% As with the {questions in the} previous section, dataset creators
% should provide answers to these questions prior to distributing the
% dataset. The questions {in this section} are intended to
% encourage dataset creators to plan for dataset maintenance and
% communicate this plan {to} dataset consumers.\\

\begin{itemize}

\item \textbf{Who {will be} supporting/hosting/maintaining the dataset?}

Institute of Automation, Chinese Academy of Sciences and Meituan Inc. will maintain \methodname{} dataset.
\item \textbf{How can the owner/curator/manager of the dataset be contacted (e.g., email address)?}

The main maintainer Yuqi Wang's e-mail: wangyuqi2020@ia.ac.cn.
\item \textbf{Is there an erratum?} If so, please provide a link or other access point.

There is no erratum for our initial release. Errata will be documented as future releases on the
dataset website.

\item \textbf{Will the dataset be updated (e.g., to correct labeling
    errors, add new instances, delete instances)?} If so, please
  describe how often, by whom, and how updates will be communicated to
  {dataset consumers} (e.g., mailing list, GitHub)?

Yes. We will update the \methodname{} dataset. Especially, we 
 will adapt to end-to-end autonomous driving tasks in the future. The update will be released on the website and GitHub.
\item \textbf{If the dataset relates to people, are there applicable
    limits on the retention of the data associated with the instances
    (e.g., were {the} individuals in question told that their data would
    {be} retained for a fixed period of time and then deleted)?} If so,
    please describe these limits and explain how they will be
    enforced.

N/A.

\item \textbf{Will older versions of the dataset continue to be
    supported/hosted/maintained?} If so, please describe how. If not,
  please describe how its obsolescence will be communicated to {dataset
  consumers}.

Yes. We will maintain the older versions of the dataset on the website and GitHub.
\item \textbf{If others want to extend/augment/build on/contribute to
    the dataset, is there a mechanism for them to do so?} If so,
  please provide a description. Will these contributions be
  validated/verified? If so, please describe how. If not, why not? Is
  there a process for communicating/distributing these contributions
  to {dataset consumers}? If so, please provide a description.

Yes. The dataset is open source under the CC BY-NC 4.0 license. So it is open to other contributors.
\item \textbf{Any other comments?}

No.
\end{itemize}

\end{document}